%% file: main.tex
\documentclass[10pt,twocolumn,letterpaper]{article}

\usepackage{cvpr}      
\usepackage{listings}
\usepackage{xcolor} 
\usepackage{multirow}
\lstdefinestyle{cvprpython}{
  language=Python,
  basicstyle=\ttfamily\footnotesize, 
  numbers=none,
  breaklines=true,            
  breakatwhitespace=true,     
  columns=fullflexible,       
  keepspaces=true,            
  showstringspaces=false,
  tabsize=2,
  xleftmargin=0pt,
  xrightmargin=0pt,
  frame=none
}
\newcommand{\ourmethod}{REMIND }
\newcommand{\emp}[1]{\textbf{\textit{#1}}}

\input{preamble}
\definecolor{cvprblue}{rgb}{0.21,0.49,0.74}
\usepackage[pagebackref,breaklinks,colorlinks,allcolors=cvprblue]{hyperref}
\definecolor{textgreen}{RGB}{57, 172, 57}
\definecolor{textred}{RGB}{200, 10, 10}
\usepackage{pifont} 
\newcommand{\cmark}{\textcolor{textgreen}{\ding{51}}}%
\def\paperID{Anonymous} 
\def\confName{CVPR}
\def\confYear{2026}

\title{\emph{REMIND:} Rethinking Medical High-Modality Learning under Missingness \\ —A Long-Tailed Distribution Perspective}

\author{
Chenwei Wu\thanks{Equal contribution.},\quad
Zitao Shuai\footnotemark[1],\quad
Liyue Shen\\
University of Michigan, Ann Arbor\\
{\tt\small liyues@umich.edu}
}

\begin{document}
\maketitle
\input{sec/0_abstract}    
\input{sec/1_intro}
\input{sec/2_abstract}
\input{sec/3_method}
\input{sec/4_experiment}

{
    \small
    \bibliographystyle{ieeenat_fullname}
    \bibliography{main}
}

\input{sec/X_suppl}

\end{document}

%% file: sec/0_abstract.tex
\begin{abstract}
Medical multi-modal learning is critical for integrating information from a large set of diverse modalities.
However, when leveraging a high number of modalities in real clinical applications, it is often impractical to obtain full-modality observations for every patient due to data collection constraints, a problem we refer to as 'High-Modality Learning under Missingness'. In this study, we identify that such missingness inherently induces an exponential growth in possible modality combinations, followed by long-tail distributions of modality combinations due to varying modality availability.  
While prior work overlooked this critical phenomenon, we find this long-tailed distribution leads to significant underperformance on tail modality combination groups. Our empirical analysis attributes this problem to two fundamental issues: 1) gradient inconsistency, where tail groups' gradient updates diverge from the overall optimization direction; 2) concept shifts, where each modality combination requires distinct fusion functions.
To address these challenges, we propose \textbf{REMIND}, a unified framework that \textbf{RE}thinks \textbf{M}ult\textbf{I}modal lear\textbf{N}ing under high-mo\textbf{D}ality missingness from a long-tail perspective. 
Our core idea is to propose a novel group-specialized Mixture-of-Experts architecture that scalably learns group-specific multi-modal fusion functions for arbitrary modality combinations, while simultaneously leveraging a group distributionally robust optimization strategy to upweight underrepresented modality combinations. 
Extensive experiments on real-world medical datasets show that our framework consistently outperforms state-of-the-art methods, and robustly generalizes across various medical multi-modal learning applications under high-modality missingness.
\end{abstract}

%% file: sec/1_intro.tex
\section{Introduction}

\begin{figure*}
    \centering
    \includegraphics[width=\linewidth]{intro_new.png}
    \caption{\textbf{Missing modalities in high-modality multi-modal learning --  A long-tailed distribution view.} }
    \label{fig:intro}
\end{figure*}

Multi-modal learning~\cite{zhu2024vision+} has emerged as a significant problem for numerous medical applications~\cite{kline2022multimodal,acosta2022multimodal}, where integrating diverse data modalities is critical for comprehensive patient assessment and clinical decision-making. These multi-modal applications often operate in a \emph{high-modality learning} regime~\cite{liang2022high}, where the number and diversity of modalities per patient can be particularly high, ranging from medical imaging, clinical notes, to laboratory measurements.
In real-world clinical scenarios, it is hard to guarantee fully observed high-modality data for each patient sample due to practical constraints in data collection protocol, such as high financial costs and radiation exposure (\eg, PET scans), patient discomfort from invasive procedures (\eg, biopsy), or technical failures~\cite{zhang2024multimodal}. 
Consequently, practical multi-modal medical datasets are usually incomplete, characterized by numerous random missing modalities and arbitrary modality combinations~\citep{yun2024flex}.

High-modality settings under such missingness often result in long-tailed distributions of modality combinations (MCs), driven by the exponential increase in possible combinations as the number of modalities grows, alongside significant variability in the availability of individual modalities. Take the Fundus Photography, Psychological Assessment, Retina Characteristics, and Multi-modal Imaging (FPRM) dataset ~\cite{FPRM} as an example, as shown in Figure~\ref{fig:intro}(a), basic modality combinations such as Electronic Health Records (EHRs) and fundus imaging are widely available across most patients; while more complex combinations involving less accessible modalities (\eg, EHR+3D scans+Fundus) occur infrequently due to the higher cost and specialized technical skills required to acquire high-quality data. Thus, the frequency distribution of modality combinations always exhibits a pronounced ``long‑tailed'' pattern (Figure~\ref{fig:intro}(b)). Despite recent research efforts to address the multi-modal missingness through both imputation-based approaches, such as learnable modality embedding banks~\citep{han2024fusemoe, yun2024flex}, and imputation-free techniques like cross-modal knowledge distillation~\citep{wei2023mmanet}, existing methods largely overlook this critical imbalance of modality combination groups.

In this work, we revisit multi-modal learning in high-modality settings with significant missingness, through the lens of long-tailed distribution modeling. We conduct experiments on several datasets using the state-of-the-art methods ~\cite{yun2024flex,han2024fusemoe,wu2025dynamic} for high-modality learning, and observe that the model performance on long-tailed modality combinations is often inferior to that on head groups, as shown in Figure~\ref{fig:intro}(b) and Sec.~\ref{sec:results}. 
This suggests that the model is not adequately trained to perform effective fusion involving less accessible modality combinations. To unveil the underlying mechanisms, we hypothesize that this is due to the inconsistent optimization of each group's performance, and further analyze the gradient dynamics of tail groups \textit{v.s.} head groups.
As shown in Figure~\ref{fig:intro}(b), we observed a phenomenon of \emp{gradient inconsistency}, 
where long-tailed groups' gradients diverge from the overall optimization direction,
which governs the model parameters' updates. 
This underscores the need for a robust approach to learn effective multi-modal fusion across all possible modality combinations including tail groups.

Addressing the long-tailed distribution of MCs introduces unique challenges beyond classical long-tailed classification problems extensively studied in previous research~\cite{zhang2025systematic, roh2021fairbatch,shi2023re}. 
In traditional long-tailed problems, the data label distributions differ across classes, yet the learned mapping function from inputs to output labels remains consistent and applies to all classes. 
In contrast, multi-modal missingness introduces a \emp{concept shift} scenario, where each MC induces a fundamentally different fusion function due to the varying availability across input modalities. Different combinations produce unique cross-modal interactions and hold distinct significance for downstream tasks~\cite{liang2023quantifying}. For example, in ICU mortality prediction, lab tests and vital signs help capture physiological state information, but adding clinical notes may create new synergistic information by contextualizing numeric abnormalities with prior patient history, requiring a fundamentally different multi-modal fusion strategy for this new modality combination.
Prior works have also shown that when certain modalities are unavailable, the models cannot easily recover the distinct modality-specific information via alignment strategies~\citep{wei2023mmanet}, and thus need to rely exclusively on signals from available sources for predictive tasks~\citep{chen2024probabilistic,huang2021makes}. 
Additionally, given the exponentially growing number of possible MCs in high-modality settings and severely limited samples in tail groups, training separate models independently for each combination is impractical. 
This challenge requires a unified framework capable of dynamically learning adaptive multi-modal fusion functions across diverse modality combinations.

To tackle the above challenges, we propose a unified multi-modal learning framework, named \emp{REMIND}, by \emp{RE}thinking \emp{M}ult\emp{I}modal lear\emp{N}ing under high-mo\emp{D}ality missingness from a long-tailed distribution perspective. 
Our approach consists of two primary components. First, we employ a group Distributionally Robust Optimization (DRO) strategy to up-weight modality combinations that are typically under-optimized due to limited data availability, and thus ensure robust performance even for rare modality combinations. Second, to accommodate the inherent concept shifts caused by modality missingness, we introduce a scalable architecture building on soft Mixture-of-Experts (MoE). It operates on a shared set of expert modules across all MC groups, but learns group-specific routing matrices that dynamically determine how to combine these experts based on the available modality combination.
Our contribution lies three-fold:

\begin{itemize}
    \item We are the first to formulate multi-modal learning with high-modality missingness through the lens of long-tailed distribution modeling, revealing that existing approaches fail on tailed modality combinations due to gradient inconsistency and concept shift challenges.
    \item We propose a novel approach combining group distributionally robust optimization with adaptive multi-modal fusion mechanisms based on soft MoE structure, enabling learning group-specific fusion functions to effectively handle imbalanced distributions of modality combinations.
    \item Extensive experiments on various real-world medical multi-modal datasets with significant missing data consistently demonstrate substantial improvements over state-of-the-art methods, particularly on challenging tailed modality combinations, validating the effectiveness of our approach.
  
\end{itemize}

%% file: sec/2_abstract.tex
\section{Related Work}
\subsection{High-Modality Learning with Missing Data}
One branch of multi-modal learning methods leverages imputation~\citep{zhang2024multimodal} to handle the missingness. 
For example, ~\citet{chen2024modality} trained a generative model to synthesize missing medical images to create full-modality records;
~\citet{liu2023m3ae, hamghalam2021modality, yun2024flex} rely on imputing the latent representations of the missing modality, with the supervision signals from prediction tasks. 
However, due to the complexity of reconstruction learning objectives, these methods either introduce instability and noise that adversely affect the primary tasks, or suffer from limited flexibility by requiring separate networks to be trained for each missing modality case~\citep{chen2024probabilistic}. 
These disadvantages become increasingly severe in the high-modality scenarios. Other imputation-free~\citep{wu2024deep} approaches aim to learn unified models that robustly extract task-relevant information from a large range of modality combinations.
\citet{liu2021incomplete,liang2019learning} apply statistical constraints to learn representation spaces that are robust to possible missing modalities. 
However, these methods still struggle to generalize to scenarios involving more than two modalities~\citep{zhao2024deep}. 
Recent works~\citep{yun2024flex,wu2025dynamic} focus on learning dynamical multi-modal fusion (\eg via MoE) functions or modeling the dynamic relationship across modalities and tasks~\citep{han2022trusted}. 
However, these methods overlook the under-optimization of samples from rare modality combinations and struggle with high-modality learning under missing data, where many modality combinations have only a few samples.

\subsection{Long-Tailed Modeling}
Long-tail learning aims to provide high precision for tail classes that have rare occurrences~\citep{shi2023re,menon2020long}, or ensure tail groups have comparable performance with the majority groups of the dataset~\citep{zhang2025systematic,liu2019large}. Classical long-tail learning methods focus on reweighting-based methods like resampling~\citep{roh2021fairbatch}, logits-adjustment~\citep{menon2020long}, distributionally robust optimization~\citep{shankar2020evaluating}, or mixing up tail groups and head groups to ensure fair optimization~\citep{chuang2021fair}. 
However, these methods share an assumption that the conditional probability $P(Y|X)$ is consistent across groups~\citep{zhang2023deep}, which does not hold in the missing modality cases. Thus, directly adapting these methods would fail to solve the high-modality learning under missingness problem.

%% file: sec/3_method.tex
\section{Method}
\subsection{Problem Formulation and Modeling}
\label{sec:problem_formulation}
\textbf{Problem Formulation.} Suppose we are given a multi-modal dataset with $m$ modalities $M_1, M_2,...,M_m$ and a prediction task $Y$. 
In practical applications, these modalities are not always completely available across different individual samples. 
This potentially yields at most $2^m-1$ missing modality cases, forming a \textbf{modality combination} family consisting of different groups $\mathcal{G}=\{\{M_1,\}, \{M_1, M_2\},...,\{M_1, M_2,...,M_m\}\}$. 
Given such a high-modality dataset with random missingness, our goal is to develop a unified multi-modal learning framework
that can accurately predict $Y$ for any arbitrary modality combination within $\mathcal{G}$~\citep{liang2022high}.

\noindent \textbf{Long-Tailed Modality Combination Groups.} 
First of all, the targeted high-modality setting involves a significantly larger number of modalities compared to classical two-modal learning scenarios (e.g., medical image and text), inherently leading to an exponential increase in possible modality patterns. For simplicity and the purpose of illustration, given a multi-modal dataset with $m$ modalities $M_1, M_2, \dots, M_m$, we assume each modality $M_i$ is independently missing with probability $p_i$. The probability of observing a specific combination $g_k$ is given by:
\[
\Pr(g_k) = \prod_{M_i \in g_k}(1 - p_i)\;\prod_{M_j \notin g_k}p_j.
\]
In practice, even slight deviations from uniform missingness probabilities (i.e., $p_i \neq 0.5$) result in the majority of samples clustering into a few dominant modality combination groups, leaving the rest as rare ``tail groups''. 
Our results in Sec.~\ref{sec:results} empirically show that directly training models on such imbalanced dataset leads to inconsistent performance across different modality combinations.
To understand this, recent studies~\citep{dong2024once, fort2020deep} suggest that the parameter updates in deep networks are largely determined by the gradient direction aligned with the principal eigenvector of the Neural Tangent Kernel (NTK) matrix, which captures the dominant gradient descent direction across the input space. 
Motivated by this, we introduce a gradient-based analysis to understand why the performance on tail modality combination groups is significantly lower. Specifically, for each group $g_k \in \mathcal{G}$, we randomly sample an equal number of data points $X_{k}$ and obtain the averaged gradient direction $GD(X^{\theta}_{k})$ for the fusion block parameterized by $\theta$. Details of gradient direction calculation are attached in the Appendix. For the whole dataset, we similarly compute the gradient direction $GD(X^{\theta})$. The gradient consistency score between group $g_k$ and the entire dataset is defined as the similarity between their gradient directions:
\begin{equation}
    sim_k = \frac{GD(X^{\theta}) \cdot GD(X^{\theta}_{k})}{||GD(X^{\theta})|| \, ||GD(X^{\theta}_{k})||}.
\end{equation}
As shown in Fig.~\ref{fig:intro}(b), for existing methods, the gradient direction of the whole dataset aligns closely with that of majority groups, maintaining high consistency scores during training. In contrast, the gradient direction for tail groups increasingly diverges from that of the whole dataset as training progresses, a phenomenon that prior research has linked to suboptimal learning outcomes~\citep{yu2020gradient, shi2023recon}. There is a clear correlation between the prevalence of modality combination groups and their gradient consistency scores, explaining impaired performance on tail groups. This phenomenon motivates the development of methods that effectively leverage each group $g_k \in \mathcal{G}$ during training, ensuring adequate optimization for samples from underrepresented groups.

\begin{figure*}[htbp]
    \centering
    \includegraphics[width=\linewidth]{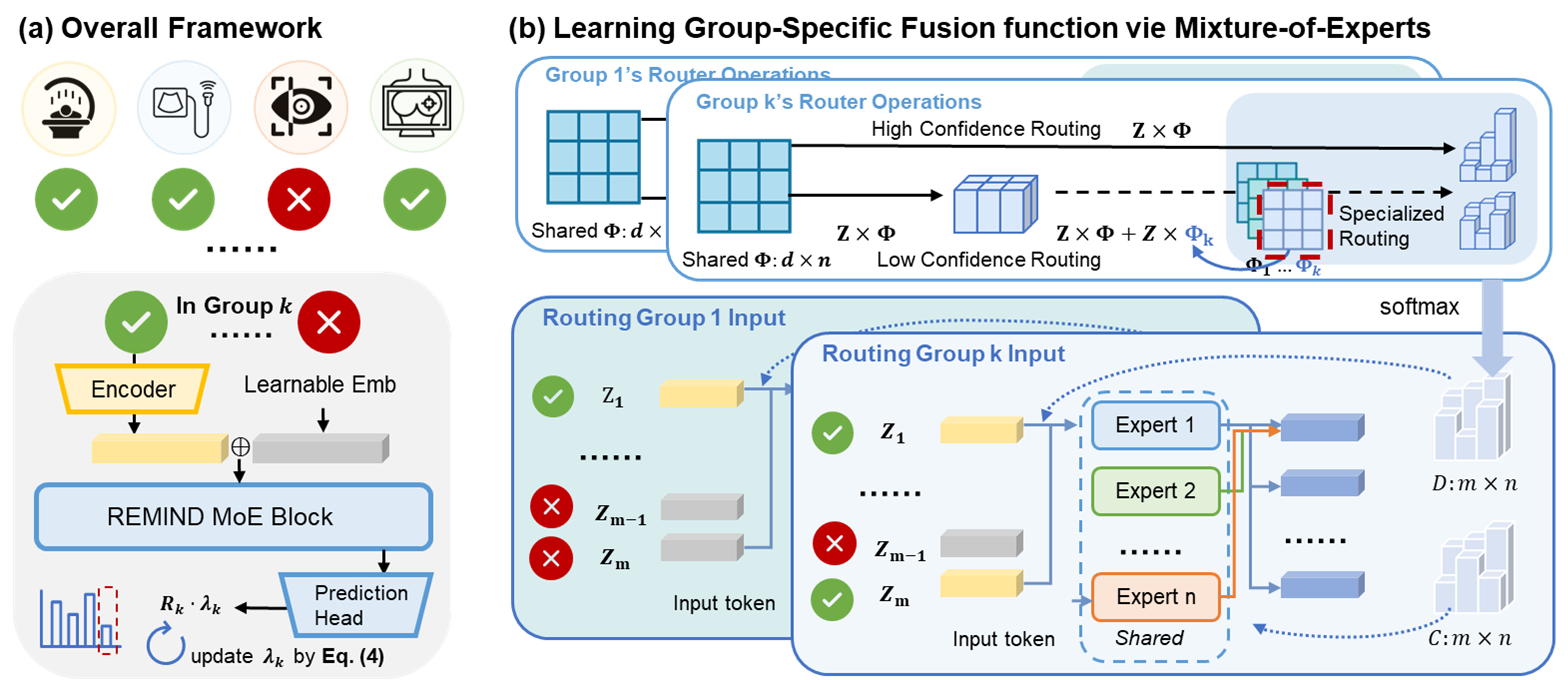}

        \caption{\textbf{Overview of REMIND.} We model high-modality learning under missing modalities.}
    \label{fig:method}
\end{figure*}

Due to the information asymmetry~\citep{chen2024probabilistic,huang2021makes} introduced by modality missingness and the unique cross-modal interactions associated with specific modality combinations~\cite{liang2023quantifying}, each modality-combination group needs a distinct and tailored fusion strategy based on the available modalities.
This requires a unified model parameterized by $\theta$ that can dynamically instantiate group-specific fusion functions $f_{g_k}(\mathbf{X}{g_k}; \theta) \rightarrow Y$ for each modality combination $g_k$. It is challenging to learn these $|\mathcal{G}|$ distinct fusion functions within a single framework while maintaining parameter efficiency across the exponentially large combination space. This motivates us to employ a Mixture-of-Experts (MoE) architecture, naturally suited for such adaptive parameter sharing and dynamic fusion learning.

\subsection{Overall Distributionally Robust Framework}
In this section, we will present the overall design of our proposed approach for learning a robust multi-modal model on datasets with long-tailed modality combination distributions.
As demonstrated in Fig.~\ref{fig:method}(a), our framework consists of a set of modality-specific feature encoders and a MoE-based fusion block for integrating multi-modal information. 
The feature encoders are first jointly optimized with the fusion block for several iterations to adapt to each modality, and then they are kept frozen in subsequent model training. 

Given a multi-modal input data sampled from arbitrary modality combination $g_k\in \mathcal{G}, k\in\{1,...,2^m-1\}$ ($|\mathcal G|=2^m-1$ is the total number of all potentially possible modality combination groups in $\mathcal{G}$), the model first extracts embeddings $Z_{i} \in \mathbb{R}^{l_i \times d}, i\in\{1,...,m\}, M_i\in g_k, $ for each available modality respectively, using the modality-specific encoders. 
Here, $d$ is the dimension of the embedding space, and $l_i$ is the token length of the embedding. 
For missing modalities, instead of zero-padding, we follow the processing in~\citet{yun2024flex} to create a set of learnable embeddings $ {B_{k}}\in \mathbb{R}^{d}, k\in\{1,...,2^m-1\}$ for each modality combination group respectively. 
For all missing modalities $M_{i'}\notin g_k$, we assign $Z_{i'}\in \mathbb{R}^{l_i \times d}$ by broadcasting $B_{k}$ to represent a group-specific yet modality-agnostic learnable embedding for the missing data within this group. 
These two sets of embeddings form a consistent input embedding space $Z_1, ..., Z_m$ shared by all the data samples. 
They are then concatenated and fed into the soft MoE block to fuse the multi-modal representations $H$, which will be projected by the task head to predict the final task output $Y$. 

As demonstrated in Sec.~\ref{sec:problem_formulation}, modality combinations follow a long‑tailed distribution and could cause inconsistent optimization across groups. To mitigate this imbalance issue, we build our approach upon the group distributionally robust optimization (DRO) framework~\citep{rahimian2019distributionally,levy2020large}. 
During training, it up‑weights samples from under‑optimized modality‑combination groups, facilitating the model to achieve more accurate prediction especially on the worst-performing groups. 
Specifically, DRO assumes the overall data can be grouped by $\mathcal{G}$ and defines a set of distributions $\mathcal{Q}^{\mathcal{G}}$ based on $\mathcal{G}$. It evaluates the model on $ \mathcal{Q}^{\mathcal{G}}$, so as to dynamically assign weights to sample from different modality combination groups (Fig.~\ref{fig:method}(a)). 
In our paper, $\mathcal{G}$ is the modality combination family. $\mathcal{Q}^\mathcal{G}$ is built on grouped distributions $D_k, k \in \{1, \dots, 2^m-1\}$ of each modality combination, and thus contains a set of potential testing domains $Q\in \mathcal{Q}^{\mathcal{G}}$. 
Formally, it is defined as \(f\)-divergence ball
\(\{Q:D_f(Q\,\|\,D^{\mathcal G})\le\rho\}\), where $D^{\mathcal G}$ is the distribution when the entire data distribution is grouped based on $\mathcal{G}$~\citep{zhang2024unified}. Here the uncertainty radius is set to zero, because we do not consider out‑of‑distribution shifts in this paper. This set of potential test domains could be further written as a convex hull $\mathcal{Q}^{\mathcal G}
=\Bigl\{\;\textstyle\sum_{k=1}^{|\mathcal G|}\lambda_k D_k
\;\bigl|\;
\lambda\in\Delta_{|\mathcal G|-1}\Bigr\}$, where \(\Delta_{|\mathcal G|-1}\) is the probability simplex.

DRO aims to consistently optimize the model's performance on each test domain $Q\in \mathcal{Q}^{\mathcal{G}}$, which is obtained by guaranteeing the model has a lower loss even in the worst-performing case. Motivated by~\citet{sagawa2019distributionally,zhang2024unified}, we adopt the simplex‑constrained formulation to optimize on the worst-case distribution: 
\begin{equation}
\footnotesize
\label{eq:dro_simplex}
\min_{\theta}\;
\max_{\lambda\in\Delta_{|\mathcal G|-1}}
\sum_{k=1}^{|\mathcal G|}\lambda_k\,R_k(\theta),
\quad
R_k(\theta)=\mathbb{E}_{(x,y)\sim D_k}\!\bigl[\ell(f_\theta(x),y)\bigr],
\end{equation}
where $\theta$ is the model parameter, $R_k(\theta)$ is the loss on the grouped distribution $D_k$. During training, we alternate between optimizing model parameters and group weights:
\begin{equation}
\footnotesize
\theta^{(t+1)}
=\theta^{(t)}
-\eta\sum_{k}\lambda_k^{(t)}\nabla_\theta R_k\bigl(\theta^{(t)}\bigr),
\end{equation}
\begin{equation}
\footnotesize
\lambda_k^{(t+1)}
=
\frac{\lambda_k^{(t)}\exp\!\bigl(\gamma\,R_k(\theta^{(t)})\bigr)}{\sum_{j}\lambda_j^{(t)}\exp\!\bigl(\gamma\,R_j(\theta^{(t)})\bigr)},
\end{equation}
where \(\eta\) is the learning rate and \(\gamma>0\) is the hyperparameter that controls the sharpness. The exponentiated update smoothly approximates \(\max_k R_k(\theta)\) while keeping non‑zero gradients for all groups~\citep{samuel2021distributional}. We refresh \(\lambda\) every \(N\) training steps in practice.

\subsection{Tackling Concept Shift with Soft MoE}
\label{sec moe}

As aforementioned, each modality combination group requires its own dynamic fusion function due to the inherent concept shift across different modality availability patterns. Mixture-of-Experts (MoE) architectures provide a natural foundation to satisfy this need, as they can dynamically create input-dependent routing pathways that have the potential to enable learning group-specific fusion strategies.

\noindent \textbf{Soft MoE-based Multi-modal Fusion Architecture.} We adopt a Soft MoE~\citep{puigcerver2023softmoe} model architecture consisting of three primary components: a self-attention module, an expert layer comprising $n$ expert networks, and a learnable routing matrix $\mathbf{\Phi}$. 
Given the modality-specific embeddings $\{Z_1,\dots,Z_m\}\in\mathbb{R}^{l\times d}$ with $l$ token length for each modality in latent dimension $d$, they are first concatenated and projected through the self-attention module to produce $\mathbf{Z}\in\mathbb{R}^{M\times d}$, where $M=l\,m$.

The router matrix $\mathbf{\Phi}\in\mathbb{R}^{d\times n}$ generates sample-adaptive routing strategies through a softmax-based assignment, providing dispatch weights $\mathbf{D}\in\mathbb{R}^{M\times n}$ and combine weights $\mathbf{C}\in\mathbb{R}^{M\times n}$. Dispatch weights $\mathbf{D}$ are computed by column-wise softmax on $\mathbf{Z}\mathbf{\Phi}$, specifically:
\begin{equation}
    D_{ij}= \frac{\exp((\mathbf{Z}\mathbf{\Phi})_{ij})}{\sum_{i'=1}^{M}\exp((\mathbf{Z}\mathbf{\Phi})_{i'j})},
\end{equation}
This dispatch weights $\mathbf{D}$ determine the modality-to-experts routing function to derive expert-specific input representations $\tilde{\mathbf{Z}}=\mathbf{D}^T\mathbf{Z}$. Each expert $j$ processes its input through a neural network $f_j$, yielding outputs $\tilde{\mathbf{Y}}_{.j}=f_j(\tilde{\mathbf{Z}}_{.j})$. Similarly, combine weights $\mathbf{C}$ are computed via a row-wise softmax:
\begin{equation}
    C_{ij}= \frac{\exp((\mathbf{Z}\mathbf{\Phi})_{ij})}{\sum_{j'=1}^{n}\exp((\mathbf{Z}\mathbf{\Phi})_{ij'})},
\end{equation}
which indicates how to combine the outputs from all experts to obtain the final fused output representation for prediction head: $\mathbf{Y}=\mathbf{C}\tilde{\mathbf{Y}}$.

\noindent \textbf{Group-Specific Adaptive Routing via Residual Matrices.} 
While Soft MoE can dynamically create fusion functions, the learned routing matrix $\mathbf{\Phi}_{\text{shared}}$ shared across all groups limits the model's ability to provide group-adaptive routing strategies, particularly for tail modality combination groups that are under-optimized during the training.

To address this limitation, instead of training separate experts per group, we propose a group-specific adaptive routing approach by introducing additional residual matrices $\mathbf{\Phi}_k$. 
It enables scalable adaptivity to a wide range of modality combinations, by facilitating both knowledge sharing across groups through shared routing matrix $\mathbf{\Phi}_{\text{shared}}$, and the extraction of fine-grained differences via residual matrix $\mathbf{\Phi}_k$. 
Specifically, for modality combination group $g_k$, we define the final group-specific adaptive routing function as:
\begin{equation}
    \mathbf{\Phi} = \mathbf{\Phi}_{\text{shared}} + \mathbf{\Phi}_k.
\end{equation}
During training, the residual design of the zero-initialized group-specific routing matrices $\mathbf{\Phi}_k$ allows them to incrementally refine rather than override the shared knowledge encoded in $\mathbf{\Phi}$. 
This ensures controlled specialization of the routing function.

Additionally, to determine when these group-specific residual adjustments should be applied, we employ an uncertainty-based gating strategy. We use the entropy of routing logits $\mathbf{Z}\mathbf{\Phi}_{\text{shared}}$ as one such uncertainty metric, though it can also be substituted by other measures (\eg, KL-divergence). 
During the model inference, we use shared matrix $\Phi_{\text{shared}}$ alone, if the entropy $H(\mathbf{Z}\mathbf{\Phi}_{\text{shared}})$ is below a predefined threshold $\xi$, which indicates confident routing. 
\begin{equation}
    H\bigl(\mathbf{Z}\mathbf{\Phi}\bigr) = -\sum_{i,j} p_{ij}\log p_{ij},
  \quad
  p_{ij} = \frac{\exp\!\bigl[(\mathbf{Z}\mathbf{\Phi})_{ij}\bigr]}
               {\sum_{i',j'}\exp\!\bigl[(\mathbf{Z}\mathbf{\Phi})_{i'j'}\bigr]}.
\end{equation}
When the uncertainty exceeds the threshold, implying uncertain modality-to-expert assignment or near-uniform routing, which are found to be suboptimal in previous MoE literature~\citep{puigcerver2023sparse,muqeeth2023soft}, the model will activate the group-specific routing adjustment $\Phi_k$ to learn a group-adaptive function. 
This mechanism preserves the model's confident routing patterns, while improving performance in cases where the shared routing matrix struggles. 

Dispatch and combine matrices are then recomputed according to the learned adapted logits $\mathbf{Z}(\mathbf{\Phi}_{\text{shared}}+\mathbf{\Phi}_k)$. Importantly, this design is highly scalable to large numbers of modalities. Unlike previous MoE designs~\cite{han2024fusemoe,wu2025dynamic,yun2024flex} that require additional modality-specific or MC-specific experts when scaling, the experts in our approach are shared across all modality combinations, and the parameter overhead is merely confined to lightweight residual routing matrices of MC groups. The choice of the threshold value and computational costs analysis are included in the Appendix.

%% file: sec/4_experiment.tex
\section{Experiments}
\begin{table*}[!htb]
\centering

\resizebox{\textwidth}{!}{%
\setlength{\tabcolsep}{8pt}
\begin{tabular}{ccccccccccccc}

\toprule[1.8pt]
\multicolumn{4}{c}{\textbf{Modality Availability}} & \textbf{Tail} &
\multicolumn{3}{c}{\textbf{Multi-modal Learning Methods}} &
\multicolumn{4}{c}{\textbf{Long-Tailed Learning Methods}} &
\textbf{Ours} \\
\cmidrule(lr){1-4}\cmidrule(lr){6-8}\cmidrule(lr){9-12}\cmidrule(lr){13-13}
$M_1$ & $M_2$ & $M_3$ & $M_4$ &  & Trans & FuseMoE & FlexMoE & FairMixup &
Reweigh & GroupDRO & FairBatch & REMIND \\ 
\midrule
\midrule
 &  & \cmark  &  & $\checkmark$ & 74.1    & 73.9    & 73.1     & 72.8      & 69.8    & 72.0      & \textbf{75.8}      & \underline{74.6}   \\
\cmark &  & \cmark &  & $\checkmark$& 78.1    & 75.3    & 79.2     & \textbf{84.2}      & 74.9    & 77.0      & \underline{82.8}      & 79.9\\
 &  &  & \cmark & $\checkmark$& 70.7    & 73.4    & 70.7     & 74.4      & 72.8    & 69.9      & \underline{74.6}      & \textbf{74.6} \\
 &  & \cmark & \cmark & & 77.2    & 76.9    & 77.5     & \underline{78.7}      & 77.7    & 78.0      & 78.0      & \textbf{79.5} \\
\cmark &  & \cmark & \cmark & $\checkmark$& \underline{73.3}    & 67.3    & 70.9     & 71.5      & 67.3    & 69.7      & 64.9      & \textbf{74.0}  \\
\cmark & \cmark & \cmark & \cmark & & 81.8    & 81.4    & 81.9     & \underline{83.2}      & 81.7    & 81.8      & 83.1      & \textbf{84.0} \\ 
\midrule
\multicolumn{5}{c}{Entire Dataset} & 78.5    & 78.2    & 78.6     & \underline{79.9}      & 78.9    & 78.9      & 79.7      & \textbf{80.7}  \\
\bottomrule[1.8pt]
\end{tabular}}
\caption{\textbf{Performance comparison (ACC$^\uparrow$) on EMBED dataset.} Across different missing-modality scenarios, REMIND achieves the overall best performance.}
\label{res:embed}
\end{table*}

\begin{table*}[!htb]
  \centering

  \resizebox{\textwidth}{!}{%
    \setlength{\tabcolsep}{9pt}
    \begin{tabular}{cccccccccccc}
    \toprule[1.8pt]
    \multicolumn{3}{c}{\textbf{Modality Availability}} & \textbf{Tail} &
    \multicolumn{3}{c}{\textbf{Multi-modal Learning Methods}} &
    \multicolumn{4}{c}{\textbf{Long-Tailed Learning Methods}} &
    \textbf{Ours} \\
    \cmidrule(lr){1-3}\cmidrule(lr){5-7}\cmidrule(lr){8-11}\cmidrule(lr){12-12}
    $M_1$ & $M_2$ & $M_3$ &  & Trans & FuseMoE & FlexMoE & FairMixup &
    Reweigh & GroupDRO & FairBatch & REMIND \\ \midrule
    \midrule
    \cmark &   &   & $\checkmark$ & 89.3 & 87.2 & 86.6 & \underline{89.3} & 86.9 & 85.2 & 86.9 & \textbf{90.8} \\
          & \cmark &   & $\checkmark$ & 80.1 & \underline{82.8} & \textbf{83.1} & 82.3 & 76.3 & 77.7 & 67.8 & 78.5 \\
    \cmark & \cmark &   &             & 84.8 & 83.4 & 84.2 & \underline{86.6} & 85.7 & 85.2 & 85.2 & \textbf{86.8} \\
          &   & \cmark & $\checkmark$ & 67.8 & 72.6 & \underline{76.2} & 67.8 & 61.5 & 64.8 & 68.6 & \textbf{76.4} \\
    \cmark &   & \cmark &             & 86.0 & 85.6 & 84.9 & 85.5 & 86.1 & \underline{86.5} & 85.1 & \textbf{88.1} \\
          & \cmark & \cmark &          & 78.8 & 79.8 & 75.6 & \textbf{82.6} & 71.8 & 81.3 & 76.4 & \underline{82.3} \\
    \cmark & \cmark & \cmark &         & 84.4 & 85.6 & 85.3 & \underline{87.7} & 86.9 & 87.6 & 86.2 & \textbf{88.0} \\
    \midrule
    \multicolumn{4}{c}{Entire Dataset} & 83.0 & 83.7 & 83.1 & \underline{85.2} & 82.5 & 84.3 & 82.6 & \textbf{86.1} \\
    \bottomrule[1.8pt]
    \end{tabular}%
  }
\caption{\textbf{Performance comparison (ACC$^\uparrow$) on MIMIC-IV dataset.} Across different missing-modality scenarios, REMIND achieves the overall best performance.}
  \label{res:mimic}
\end{table*}

\begin{table*}[!htb]
  \centering

  \resizebox{\textwidth}{!}{%
    \setlength{\tabcolsep}{8pt}
    \begin{tabular}{ccccccccccccc}
    \toprule[1.8pt]
    \multicolumn{4}{c}{\textbf{Modality Availability}} & \textbf{Tail} &
    \multicolumn{3}{c}{\textbf{Multi-modal Learning Methods}} &
    \multicolumn{4}{c}{\textbf{Long-Tailed Learning Methods}} &
    \textbf{Ours} \\
    \cmidrule(lr){1-4}\cmidrule(lr){6-8}\cmidrule(lr){9-12}\cmidrule(lr){13-13}
    $M_1$ & $M_2$ & $M_3$ & $M_4$ &  & Trans & FuseMoE & FlexMoE & FairMixup &
    Reweigh & GroupDRO & FairBatch & REMIND \\ \midrule
    \midrule
    \cmark &   &   &   & $\checkmark$ & 82.1 & 84.0 & 84.3 & 83.6 & 83.2 & \underline{84.8} & 84.3 & \textbf{85.5} \\
    \cmark & \cmark & \cmark &   & $\checkmark$ & \underline{89.0} & 85.2 & 74.8 & 76.8 & 86.1 & 84.2 & 82.9 & \textbf{93.8} \\
    \cmark &   &   & \cmark &            & 81.4 & 81.5 & \underline{82.3} & 78.8 & 81.4 & 82.2 & 82.0 & \textbf{82.5} \\
    \cmark & \cmark &   & \cmark & $\checkmark$ & 66.7 & 66.7 & 75.0 & 78.3 & 66.7 & \underline{83.3} & 75.0 & \textbf{100} \\
    \cmark & \cmark & \cmark & \cmark & $\checkmark$ & 77.6 & \underline{79.7} & 74.8 & 71.1 & 76.0 & 73.2 & 77.0 & \textbf{86.2} \\
    \midrule
    \multicolumn{5}{c}{Entire Dataset} & 81.0 & \underline{81.5} & 80.9 & 78.2 & 80.8 & 81.2 & 81.4 & \textbf{83.8} \\
    \bottomrule[1.8pt]
    \end{tabular}%
  }
  \caption{\textbf{Performance comparison (ACC$^\uparrow$) on FPRM dataset.} Across different missing-modality scenarios, REMIND achieves the overall best performance.}
  \label{res:fprm}
  \vspace{-10pt}
\end{table*}

\noindent \textbf{Datasets.} To verify the generalizability of \ourmethod  across high-modality multi-modal learning applications, we conduct extensive experiments over three publicly available medical multi-modal datasets for different tasks. 
Each dataset consists of at least three modalities with significant missing modality issues. Specifically, we use:
\begin{itemize}
    \item EMBED (EMory BrEast Imaging Dataset)~\citep{jeong2023emory} contains 4 modalities, including M1: C-View synthetic Cranio-Caudal view (CC), M2: C-View synthetic 2-D Medio-Lateral Oblique view (MLO), M3: full-field digital mammography (FFDM) CC, and M4: FFDM MLO, which are used to train a breast density prediction model. 
    \item MIMIC-IV (Medical Information Mart for Intensive Care IV Dataset)~\citep{johnson2023mimic}. Following~\cite{han2024fusemoe}, we extract 3 modalities of M1: ICD-9 codes, M2: clinical text, and M3: labs' vital values, and use them to perform the 48-hour mortality prediction task. 
    \item FPRM (Multi-modal Eye Imaging, Retina Characteristics, and Psychological Assessment Dataset)~\citep{zhang2024multimodal}, where we utilize 4 modalities, including M1: 2D fundus photos, M2: 3D videos of retinal blood flow, M3: 2D oxygen saturation, and M4: tabular data for health deterioration grade classification. For evaluation, we report the accuracy and F1-score of the compared methods on each classification task. 
\end{itemize}
In all experiments, we run each method with 3 different random seeds and report the mean and standard deviation of the performance. 
Groups with approximately less than 15\% frequency are denoted 'Tail groups'. Full results are in the Appendix.

\noindent \textbf{Baselines.} We compare \ourmethod with the state-of-the-art methods in tackling missingness in high-modality learning, including Soft MoE~\citep{puigcerver2023sparse,wu2025dynamic}, FuseMoE~\citep{han2024fusemoe}, FlexMoE~\citep{yun2024flex}.
Besides, we also compare with other long-tailed learning approaches, by adapting them to multi-modal learning models. 
Specifically, we combine multi-modal MoE with group robustness strategies including Distribution Adjustment~\citep{menon2020long}, GroupDRO~\citep{shankar2020evaluating}, FairBatch~\citep{roh2021fairbatch}, and FairMixup~\citep{chuang2021fair}.

\noindent \textbf{Implementation Details.} We use ViT-Base~\citep{dosovitskiy2020image} as the feature encoder for image modalities in EMBED, and the T5 encoder~\citep{raffel2020exploring} for text modalities in the MIMIC-IV dataset. For the FPRM dataset, we use a 3D-ResNet~\citep{tran2018closer} for 3D modalities, 2D-ResNet~\citep{he2016deep} encoders to encode 2D image modalities, and a patch embedder to encode tabular modalities. In main experiments, we set the batch size to $32$, and use the Adam optimizer~\citep{kingma2014adam} with a weight decay of $5e-5$. For EMBED and FPRM datasets, 
we use the learning rate $5e-5$, and $1e-4$ learning rate for the MIMIC dataset. 
More implementation details and hyperparameter ablation analysis are included in the Appendix.

\section{Results}
\label{sec:results}
\subsection{Main Results}
 \noindent \textbf{Our method better handles high-modality learning under missingness.} As shown in Tables~\ref{res:embed},~\ref{res:mimic}, and~\ref{res:fprm}, our method demonstrates consistent superior performance across various datasets, under different missing modality scenarios.
 Either simply applying existing group robustness methods to multi-modal learning, or directly fusing multi-modalities, fail to address the core challenges posed by long‑tailed modality distributions.

\noindent \textbf{Our method excels on tail groups,} where existing state‑of‑the‑art methods such as FuseMoE, FlexMoE, and Soft MoE struggle. 
Our approach achieves superior accuracy regardless of whether tail groups consist of single modalities (MIMIC) or complex/full modality combinations (FPRM).

\subsection{Analysis}

\begin{figure}
    \centering
    \includegraphics[width=\linewidth]{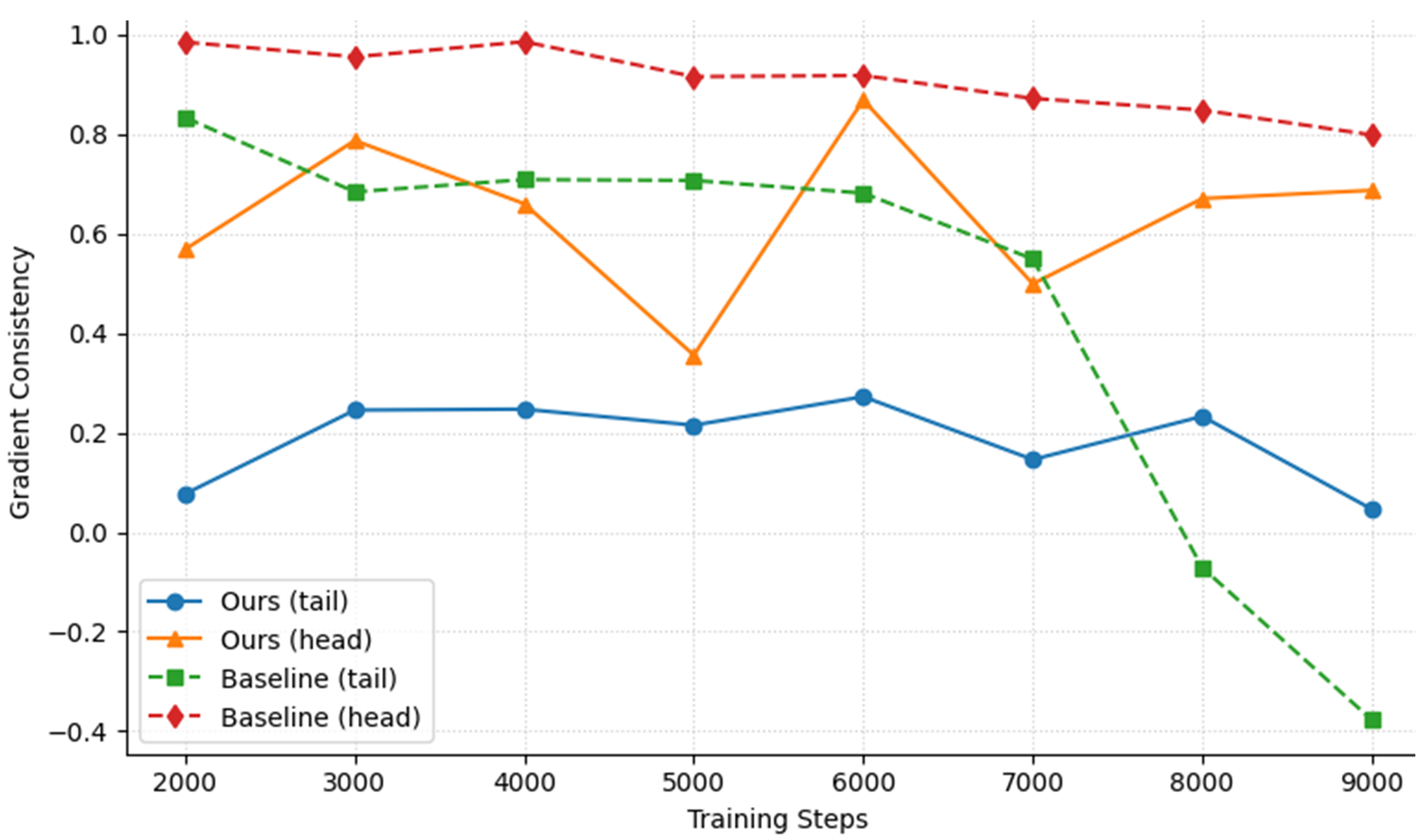}
    \caption{\textbf{Gradient inconsistencies across training steps.}}
    \label{fig:grad-resolve}
\end{figure}

\noindent \textbf{Q1: Has gradient inconsistency been alleviated?} Our earlier analysis identified a significant correlation between gradient consistency and the frequency of modality combinations encountered in datasets. Specifically, rare modality combinations tend to exhibit substantial gradient divergence, severely limiting model learning. Figure~\ref{fig:grad-resolve} demonstrates that this gradient inconsistency problem worsens as training progresses. The head groups maintain high gradient consistency throughout training, creating a stark divergence that prevents effective learning on tail combinations. In contrast, our method demonstrates a more stable gradient consistency for both head and tail groups throughout the training.
REMIND helps global updates become less head-dominated; head-vs-global consistency drops while the head–tail gap shrinks. Ablation results (Table~\ref{tab:ablation}) further verify that both components of DRO and expert specialization are crucial for alleviating this issue and improving model performance.

\begin{table}[t]
\small
\centering

\resizebox{\columnwidth}{!}{%
\begin{tabular}{ccccccc}
\toprule[1.8pt]
\textbf{M1} & \textbf{M2} & \textbf{M3} & \textbf{M4} & \textbf{W/o DRO} & \textbf{W/o Expert Spec.} & \textbf{Ours} \\
\midrule
\midrule
  &   & 1 &   & 73.6 & 72.0 & 74.6 \\
1 &   & 1 &   & 80.6 & 82.8 & 79.9 \\
  &   &   & 1 & 75.1 & 68.8 & 74.6 \\
  & 1 & 1 &   & 79.1 & 78.7 & 79.5 \\
1 &   & 1 & 1 & 72.7 & 72.7 & 74.0 \\
1 & 1 & 1 & 1 & 84.3 & 83.8 & 84.0 \\
\midrule
\multicolumn{4}{c}{Entire Dataset}   & 80.5 & 80.0 & 80.7 \\
\bottomrule[1.8pt]
\end{tabular}%
}
\caption{\textbf{Ablation studies on EMBED dataset.}}
  \label{tab:ablation}
  \vspace{-16pt}
\end{table}

\begin{figure}
    \centering
    \includegraphics[width=1\linewidth]{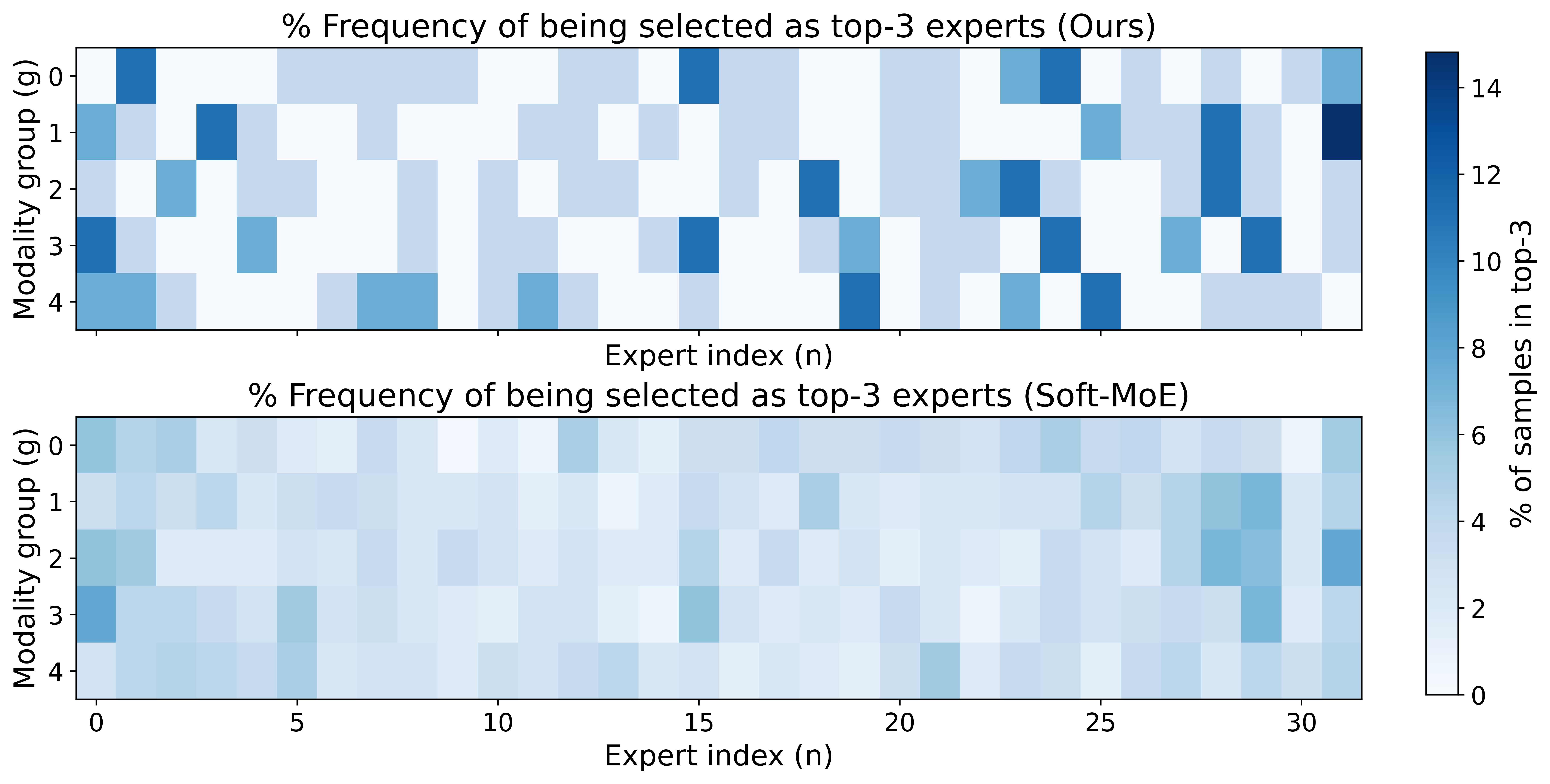}
    \caption{\textbf{Visualization of top experts patterns across modality combination groups on FPRM Dataset.}}
    \label{fig:experts_freq}
    \vspace{-10pt}
\end{figure}

\noindent\textbf{Q2: Have we achieved meaningful expert specialization?}
To investigate whether our group-adaptive MoE architecture learns meaningful specialization patterns, we visualize the heatmaps of expert assignment scores across different modality combination groups (Fig.~\ref{fig:experts_freq}). The visualization shows the frequency with which each expert falls into the top-3 experts for different modality combinations. Our method exhibits significantly clearer and more distinct specialization patterns across modality combinations compared to the baseline Soft MoE, verifying \ourmethod has facilitated learning of group-specific fusion functions.

\noindent \textbf{Q3: Can our method handle extreme missing modality scenarios?} Real clinical scenarios often involve certain modalities with extremely high missing rates, which raises the question of whether incorporating such sparsely available modalities in multi-modal model development provides any benefit. To investigate whether our method can handle such scenarios, we manually created extreme missing scenarios for each modality in the FPRM dataset by simulating 80\% missing rates. As shown in Table~\ref{tab:extreme_missing}, for each modality $M_i$, we start from the subset of FPRM dataset where all modalities are available, and artificially remove a random 80\% of samples from modality $M_i$, creating two distinct sample groups: $M_i$ Available and $M_i$ Absent. We then evaluate overall performance and performance on each group separately to assess whether methods can (1) effectively utilize sparse modalities when available and (2) maintain robustness when they are absent. Results demonstrate that even under extreme missingness, our method can still effectively incorporate these sparse modalities to achieve substantial performance improvements over baselines.

\begin{table}[t]
\small
\centering

\resizebox{\columnwidth}{!}{%
\begin{tabular}{lcccccc}
\toprule[1.8pt]
& \multicolumn{3}{c}{\textbf{Basic Soft MoE}} & \multicolumn{3}{c}{\textbf{Ours}} \\
\cmidrule(r){2-4}\cmidrule(l){5-7}
& Overall & $M_i$ Available & $M_i$ Absent & Overall & $M_i$ Available & $M_i$ Absent \\
\midrule
\midrule
\textbf{M1}  & 70.7 & 81.5 & 68.1 & 78.1 (+7.4\%) & 84.3 (+2.8\%) & 76.6 (+8.5\%) \\
\textbf{M2}  & 73.9 & 71.5 & 74.6 & 84.1 (+10.2\%) & 84.2 (+12.7\%) & 84.1 (+9.5\%) \\
\textbf{M3}  & 76.4 & 76.0 & 76.5 & 79.6 (+3.2\%) & 85.3 (+9.3\%) & 78.2 (+1.7\%) \\
\textbf{M4}  & 79.4 & 79.4 & 79.4 & 87.3 (+7.9\%) & 87.8 (+8.4\%) & 87.2 (+7.8\%) \\
\bottomrule[1.8pt]
\end{tabular}%
}
\caption{\textbf{Performance under 80\% artificial missingness for each modality on FPRM dataset.} For each row, 80\% of samples have modality $M_i$ removed. REMIND brings large improvements}
\label{tab:extreme_missing}
\end{table}

\begin{figure}
    \centering
    \includegraphics[width=1\linewidth]{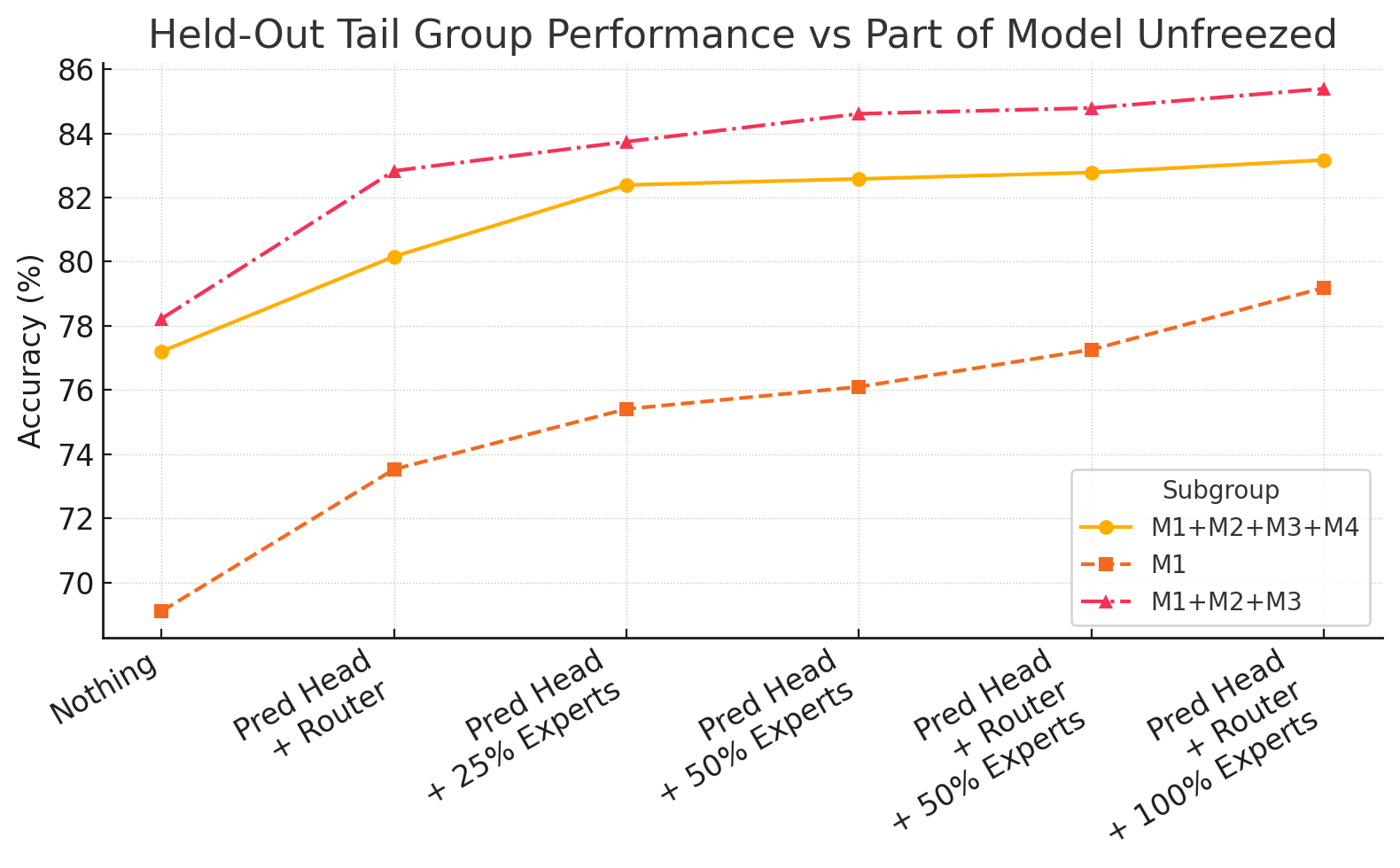}
    \vspace{-16pt}\caption{\textbf{Performance of held-out tail groups in the FPRM dataset.} We show performances when unfreezing and finetuning different model parts. X-Axis indicates the parts of model being finetuned: from zero-shot ('Nothing') to full finetuning ('Pred Head +Router + 100\% experts').}
    \label{fig:unseen}
    \vspace{-10pt}
\end{figure}

\noindent \textbf{Q4: Can our method adapt to unseen modality combinations?}
One interesting finding is that our framework can be easily adapted to modality combinations unseen during training by only finetuning the router matrix and prediction head. 
We simulated this scenario by completely excluding certain modality combinations from the training set, then evaluating different fine-tuning strategies on them.
As shown in Fig.~\ref{fig:unseen}, our analysis reveals that unfreezing the prediction head and routing matrices could already achieve good performance with minimal parameter updates. 
Our group-adaptive mechanism is able to learn transferable representations and allows core expert knowledge to be shared across groups, thus requiring only router adaptation for novel modality combinations.

\section{Conclusion}
In this paper, we introduce REMIND, a novel framework for high-modality learning under missingness. Our analysis links prior methods' underperformance on long-tailed modality combinations to gradient inconsistency and concept shifts. Extensive experiments demonstrate that REMIND consistently improves performance, particularly on tail modality-combination groups and under extreme missing-modality scenarios.

\section*{Acknowledgment}
The author acknowledges funding support by NSF (National Science Foundation) via grant IIS-2435746, Defense Advanced Research Projects Agency (DARPA) under Contract No.~HR00112520042, Hyundai America Technical Center, Inc. (HATCI), as well as the University of Michigan MICDE Catalyst Grant Award and MIDAS PODS Grant Award.

%% file: sec/X_suppl.tex
\clearpage
\section{Appendix}
\subsection{Intuition Behind Gradient Consistency Analysis}
\label{sec:gradient_consistency_intuition}

In this section, we provide an intuitive explanation of our gradient consistency analysis and why it reveals the under-optimization of tail modality combination groups.

\subsubsection{The Core Problem: Which Examples Drive Training?}

When training a neural network with gradient descent, not all examples contribute equally to parameter updates. The actual update direction is determined by the aggregate gradient across the entire dataset. If certain subgroups (e.g., tail modality combinations) have gradient directions that diverge from this aggregate direction, their optimization objectives are effectively ignored during training.

\subsubsection{From Individual Gradients to Dominant Direction}

Given a dataset $\mathcal{X} = \{x_1, \ldots, x_n\}$ and model parameters $\theta$, we can compute the gradient of each example's loss with respect to the parameters: $\nabla_\theta \ell(f(x_i; \theta), y_i)$. These gradients form the Jacobian matrix:
\begin{equation}
\mathbf{J}_\theta(\mathcal{X}) = \begin{bmatrix}
\nabla_\theta f(x_1; \theta) \\
\nabla_\theta f(x_2; \theta) \\
\vdots \\
\nabla_\theta f(x_n; \theta)
\end{bmatrix} \in \mathbb{R}^{n \times p},
\end{equation}
where $p$ is the number of parameters. Each row represents how one example's output changes with respect to all parameters.

The Neural Tangent Kernel (NTK) matrix captures how these individual gradients interact:
\begin{equation}
\mathbf{\Theta}(\mathcal{X}, \mathcal{X}) = \mathbf{J}_\theta(\mathcal{X}) \mathbf{J}_\theta(\mathcal{X})^\top \in \mathbb{R}^{n \times n},
\end{equation}
where $\Theta_{ij} = \langle \nabla_\theta f(x_i), \nabla_\theta f(x_j) \rangle$ measures the alignment between example $i$'s and example $j$'s gradients.

\subsubsection{Extracting the Dominant Gradient Direction}

The NTK matrix $\mathbf{\Theta}$ is symmetric and positive semi-definite, so we can perform eigendecomposition:
\begin{equation}
\mathbf{\Theta} = \mathbf{U} \mathbf{\Lambda} \mathbf{U}^\top = \sum_{i=1}^{n} \lambda_i \mathbf{u}_i \mathbf{u}_i^\top,
\end{equation}
where $\lambda_1 \geq \lambda_2 \geq \cdots \geq 0$ are eigenvalues and $\{\mathbf{u}_i\}$ are orthonormal eigenvectors.

The top eigenvector $\mathbf{u}_{\max} = \mathbf{u}_1$ (corresponding to the largest eigenvalue $\lambda_1$) captures the dominant direction in example-space---the linear combination of examples that collectively exerts the strongest influence on training dynamics. Projecting the Jacobian onto this direction gives us the aggregate parameter gradient:
\begin{equation}
\mathbf{g}_\theta(\mathcal{X}) = \mathbf{u}_{\max}^\top \mathbf{J}_\theta(\mathcal{X}) \in \mathbb{R}^p.
\end{equation}

\textbf{Intuition:} $\mathbf{u}_{\max}$ identifies which weighted combination of examples dominates the gradient updates. If head groups (with many samples) align with $\mathbf{u}_{\max}$, they drive training. If tail groups diverge from $\mathbf{u}_{\max}$, they are under-optimized.

\subsubsection{Measuring Gradient Consistency for Subgroups}

For a modality combination group $g_k \subseteq \mathcal{X}$, we repeat the process:
\begin{enumerate}
    \item Compute $\mathbf{J}_\theta(\mathcal{X}_{g_k})$ for samples in group $g_k$.
    \item Form $\mathbf{\Theta}(\mathcal{X}_{g_k}, \mathcal{X}_{g_k}) = \mathbf{J}_\theta(\mathcal{X}_{g_k}) \mathbf{J}_\theta(\mathcal{X}_{g_k})^\top$.
    \item Extract the top eigenvector $\mathbf{u}'_{\max}$ and compute $\mathbf{g}_\theta(\mathcal{X}_{g_k}) = (\mathbf{u}'_{\max})^\top \mathbf{J}_\theta(\mathcal{X}_{g_k})$.
\end{enumerate}

We then define the \textbf{gradient consistency score} as the cosine similarity between the group's dominant gradient direction and the overall dataset's dominant gradient direction:
\begin{equation}
\text{GC}_\theta(g_k) = \frac{\mathbf{g}_\theta(\mathcal{X}) \cdot \mathbf{g}_\theta(\mathcal{X}_{g_k})}{\|\mathbf{g}_\theta(\mathcal{X})\| \|\mathbf{g}_\theta(\mathcal{X}_{g_k})\|}.
\end{equation}

\textbf{Interpretation:}
\begin{itemize}
    \item $\text{GC}_\theta(g_k) \approx 1$: Group $g_k$'s preferred update direction aligns with the overall training direction $\Rightarrow$ group is well-optimized.
    \item $\text{GC}_\theta(g_k) \ll 1$: Group $g_k$'s preferred direction diverges from overall training $\Rightarrow$ group is under-optimized.
\end{itemize}

\subsubsection{Why This Matters for Long-Tailed Modality Combinations}

In our experiments (Figure~\ref{fig:intro}(b)), we observe:
\begin{itemize}
    \item \textbf{Head groups} (frequent modality combinations) maintain high gradient consistency throughout training because their large sample counts dominate the NTK's top eigenvector $\mathbf{u}_{\max}$.
    \item \textbf{Tail groups} (rare modality combinations) exhibit low and decreasing gradient consistency as training progresses, indicating their optimization objectives are increasingly ignored.
\end{itemize}

This gradient divergence explains the performance gap between head and tail groups and motivates our group DRO approach, which explicitly reweights groups to prevent tail groups from being overlooked during optimization.

\subsubsection{Connection to Neural Tangent Kernel Theory}

Recent work~\citep{fort2020deep,dong2024once} has shown that in the NTK regime (wide networks with small learning rates), the gradient descent dynamics are governed by the kernel's eigenspectrum. The top eigenvector $\mathbf{u}_{\max}$ determines which directions in function space are learned fastest. Our analysis leverages this insight to quantify when subgroups are aligned or misaligned with the dominant learning direction, providing a principled explanation for the long-tail performance gap.

\subsubsection*{Recap of key formulas}
\begin{align}
  \Theta(X,X) &= J_\theta(X)\,J_\theta(X)^{T}, \\
  \Theta &= U\,\Lambda\,U^{T}, 
     \quad u_{\max} = \text{eigenvector of }\lambda_1, \\
  g_\theta(X) &= u_{\max}^T\,J_\theta(X), \\
  \mathrm{GC}_\theta(X')
     &= \frac{\,g_\theta(X)\cdot g_\theta(X')}
              {\|g_\theta(X)\|\;\|g_\theta(X')\|}.
\end{align}

\subsection{Soft MoE Routing and Uncertainty in Routing}
This section describes the core Soft Mixture‐of‐Experts (SoftMoE) routing pipeline:
normalization, logits, probabilities, deep entropy analysis, uncertainty metrics (including KL divergence),
and final expert mixing. Each step shows PyTorch code, tensor shapes, mathematical formulas, and intuition.

\subsubsection{1. Input \& Shapes}
\begin{itemize}
  \item \textbf{Input embeddings:} $x \in \mathbb{R}^{B \times M \times D}$
    \begin{itemize}
      \item $B$: batch size
      \item $M$: sequence length (\# tokens)
      \item $D$: feature dimension
    \end{itemize}
\end{itemize}

\subsubsection{2. Normalize Input \& Router Weights}
\paragraph{Code (PyTorch)}
\begin{lstlisting}[language=Python]
# Normalize token embeddings
x_norm   = F.normalize(x, dim=-1)               # [B, M, D]
# phi: router params, shape [D, N, P]
phi_norm = self.scale * F.normalize(self.phi, dim=0)  # [D, N, P]
\end{lstlisting}
\paragraph{Math}
\begin{align*}
\tilde x_{b,m,:} &= \frac{x_{b,m,:}}{\|x_{b,m,:}\|_2}, \\
\tilde\phi_{:,i,p} &= s \;\frac{\phi_{:,i,p}}{\|\phi_{:,i,p}\|_2}.
\end{align*}
\paragraph{Intuition} Decouples magnitude from direction for stable routing.

\subsubsection{3. Compute Routing Logits}
\paragraph{Code}
\begin{lstlisting}[language=Python]
# Compatibility scores
logits = torch.einsum("bmd,dnp->bmnp", x_norm, phi_norm)  # [B, M, N, P]
\end{lstlisting}
\paragraph{Math}
\[
L_{b,m,i,p} = \sum_{d=1}^D \tilde x_{b,m,d}\;\tilde\phi_{d,i,p}.
\]
\paragraph{Intuition} Higher $L$ means token $m$ matches expert $i$ slot $p$ strongly.

\subsubsection{4. Convert to Probabilities (Softmax)}
\paragraph{Code}
\begin{lstlisting}[language=Python]
prob = torch.softmax(logits, dim=2)  # [B, M, N, P]
\end{lstlisting}
\paragraph{Math}
\[
p_{b,m,i,p} = \frac{\exp(L_{b,m,i,p})}{\sum_{j=1}^N \exp(L_{b,m,j,p})}.
\]

\subsubsection{5. Deep Dive: Entropy as Uncertainty}
Entropy quantifies the \emph{spread} of the router’s distribution over experts (and slots):
\paragraph{Code}
\begin{lstlisting}[language=Python]
entropy = -(prob * torch.log(prob + 1e-8)).sum(dim=2)  # [B, M, P]
\end{lstlisting}
\paragraph{Math}
\[
H_{b,m,p} = -\sum_{i=1}^N p_{b,m,i,p}\,\ln p_{b,m,i,p}, \quad 0 \le H \le \ln N.
\]
\[
\text{For }N=32,\quad H_{\max} = \ln(32) \approx 3.47\text{ nats.}
\]
\paragraph{Interpretation}
\begin{itemize}
  \item $H \to 0$ (peaky): single expert dominates $\Rightarrow$ high certainty.
  \item $H \to \ln N$ (flat): uniform distribution $\Rightarrow$ high uncertainty.
\end{itemize}

\subsubsection{5.1 Entropy in Bits vs.\ Nats}
\begin{itemize}
  \item \textbf{Nats:} uses natural log, max $=\ln N$.
  \item \textbf{Bits:} use base‐2 log, max $=\log_2 N$.
\end{itemize}
\begin{lstlisting}[language=Python]
entropy_bits = -(prob * torch.log2(prob + 1e-8)).sum(dim=2)
\end{lstlisting}

\subsubsection{5.2 Numerical Considerations}
Add $\epsilon$ inside log to avoid $\log0$, then:
\begin{itemize}
  \item Clamp: \texttt{entropy = entropy.clamp\_min(0)}.
  \item Or: \texttt{torch.where(prob>0, torch.log(prob), 0)}.
\end{itemize}

\subsubsection{6. Certainty \& Uncertainty Metrics}
We collect several metrics from $\mathrm{prob}\in\mathbb{R}^{B\times M\times N\times P}$, all shape $[B,M,P]$:
\begin{table}[h]
\small
\centering
\begin{tabular}{l|l|l}
\bf Metric           & \bf Definition                                                      & \bf Range       \\
\hline
Normalized Certainty & $1 - H/\ln N$                                                      & [0,1]          \\
Max Probability      & $\max_i p_i$                                                       & [1/N,1]        \\
Margin               & $p_{(1)} - p_{(2)}$                                                & [0,1-1/N]      \\
Gini Impurity        & $1 - \sum_i p_i^2$                                                 & [0,1-1/N]      \\
Variance             & $\sum_i (p_i - 1/N)^2$                                             & [0,?]          \\
KL vs Uniform        & $D_{KL}(p\|u)=\sum_i p_i\ln\frac{p_i}{1/N}$                        & [0,$\ln$ N]      \\
\end{tabular}
\caption{Uncertainty metrics over router probabilities}
\end{table}

\paragraph{Code Example}
\begin{lstlisting}[language=Python]
# Max-prob
max_prob, _ = prob.max(dim=2)        # [B, M, P]
# Margin
sorted_p, _  = prob.sort(dim=2, descending=True)
margin       = sorted_p[...,0] - sorted_p[...,1]
# KL divergence from uniform
u       = 1.0 / self.num_experts
kl_div  = (prob * (torch.log(prob+1e-8) - torch.log(u))).sum(dim=2)
\end{lstlisting}

\subsubsection{7. Aggregating Over Dimensions}
Aggregate $[B,M,P]$ to:
\begin{itemize}
  \item Per‐token: mean or max over slots $P$.
  \item Per‐example: mean over tokens $M$.
  \item Global: mean over batch $B$.
\end{itemize}
\begin{lstlisting}[language=Python]
cert_seq    = certainty.mean(dim=-1)      # [B]
avg_entropy = entropy.mean(dim=[1,2])     # [B]
\end{lstlisting}

\subsubsection{8. Final Expert Routing \& Mixing}
\paragraph{Code}
\begin{lstlisting}[language=Python]
# Routing weights
d = torch.softmax(logits, dim=2)                     # [B, M, N, P]
# Combine weights
c = torch.softmax(logits.flatten(2), dim=-1).view_as(d)
# Expert-slot inputs
xs = torch.einsum("bmd,bmnp->bnpd", x_norm, d)        # [B, N, P, D]
# Expert outputs
y_splits = [f(xs[:,i]) for i,f in enumerate(self.experts)]
ys       = torch.stack(y_splits, dim=1)              # [B, N, P, D]
# Merge back
y = torch.einsum("bnpd,bmnp->bmd", ys, c)             # [B, M, D]
\end{lstlisting}

\paragraph{Shapes Recap}
\begin{tabular}{l|l}
Tensor                & Shape        \\
\hline
$x_{\mathrm{norm}}$   & $[B,M,D]$    \\
$\phi_{\mathrm{norm}}$& $[D,N,P]$    \\
$\mathrm{logits}$     & $[B,M,N,P]$  \\
$\mathrm{prob}$       & $[B,M,N,P]$  \\
$\mathrm{entropy}$    & $[B,M,P]$    \\
$\mathrm{certainty}$  & $[B,M,P]$    \\
$d$                   & $[B,M,N,P]$  \\
$xs$                  & $[B,N,P,D]$  \\
$ys$                  & $[B,N,P,D]$  \\
$c$                   & $[B,M,N,P]$  \\
$y$                   & $[B,M,D]$    \\
\end{tabular}

\subsubsection{Practical Insights on Entropy and KL Thresholds in Soft MoE}

In practice, although our original paper employed an entropy-based threshold of \(0.8\times\text{Max\_Entropy}\), using a KL-divergence threshold of 0.1 yields similar performance. This equivalence arises because, without explicit regularization or expert specialization, the routing entropy naturally approaches the maximum entropy \(\ln N\) for nearly all tokens, reflecting uniformly distributed routing probabilities across experts. Consequently, both a high entropy threshold and a low KL-divergence threshold effectively identify tokens exhibiting high uncertainty (near-uniform routing).

However, after regularization and expert specialization, the routing probabilities tend to cluster around a few experts, appearing visually sparse. Despite this apparent sparsity, the actual entropy value does not significantly decrease toward zero, due to the inherently soft nature of the routing probabilities in Soft MoE. This contrasts sharply with traditional sparse MoE systems.

To illustrate, consider a numerical example with \(N=32\) experts, where probabilities primarily cluster around five experts: \([0.2, 0.2, 0.15, 0.1, 0.05]\), totaling 0.7 probability mass, with the remaining 0.3 uniformly distributed among the other 27 experts. Computing the entropy and KL divergence from uniform for this distribution yields:

\begin{align*}
H &\approx 2.658\;\text{nats},\\
H_{\max} &= \ln(32)\approx 3.466\;\text{nats},\\
D_{\mathrm{KL}}(p\|u) &\approx 0.807.
\end{align*}

Even in this significantly specialized scenario, the entropy remains notably high (around 77 per cent of the maximum), clearly demonstrating that the entropy value in Soft MoE does not approach zero despite visual sparsity. Thresholding is good but not critically important in this paper, any reasonable value (depending on dataset and modalities) around 0.8-0.95 for Entropy and around 0.6-0.15 for KL empirically works for our use case.
\subsection{Additional Results}

As discussed in the main text, we ran 3 different splits ( Random Seed 2, 42, 778 for FPRM, seed 0, 1, 42 for EMBED and MIMIC, these are chosen at random) and here are the Accuracy Standard Deviation, F1 Mean, and F1 Standard Deviation results. \textit{Readers may notice there is a MC for FPRM with 100\% accuracy and that is because  this is a very small MC (0.2 (test)* 0.8\% * 3394 samples). Results are consistent across 3 runs.}

\begin{table*}[!t]
\centering
\renewcommand{\arraystretch}{1.2}
\label{tab:fprm-f1-mean}
\begin{tabular*}{\textwidth}{@{\extracolsep{\fill}} c c c c c c c c c c c c }
\toprule[1.8pt]
\multicolumn{4}{c}{\textbf{Modality}} 
  & \multicolumn{3}{c}{\textbf{Multi‐modal learning methods}} 
  & \multicolumn{4}{c}{\textbf{Long tail}} 
  & \textbf{Ours} \\
\cmidrule(lr){1-4}\cmidrule(lr){5-7}\cmidrule(lr){8-11}\cmidrule(lr){12-12}
M1 & M2 & M3 & M4 
  & SoftMoE & FuseMoE & FlexMoE 
  & FairMixup & Reweigh & GroupDRO & FairBatch 
  & REMIND \\
\midrule
\midrule
1 & 0 & 0 & 0 
  & 0.772 & \underline{0.796} & 0.770 
  & 0.635 & 0.760 & \textbf{0.798} & 0.729 
  & 0.752 \\
1 & 1 & 1 & 0 
  & 0.870 & \underline{0.888} & 0.871 
  & 0.708 & 0.590 & 0.799 & 0.858 
  & \textbf{0.893} \\
1 & 0 & 0 & 1 
  & 0.671 & 0.758 & 0.756 
  & 0.559 & \textbf{0.784} & 0.749 & 0.648 
  & \underline{0.765} \\
1 & 1 & 0 & 1 
  & 0.521 & 0.521 & 0.711 
  & 0.587 & 0.521 & \underline{0.743} & 0.587 
  & \textbf{1.000} \\
1 & 1 & 1 & 1 
  & 0.671 & \underline{0.743} & 0.729 
  & 0.545 & 0.497 & 0.687 & 0.657 
  & \textbf{0.765} \\
\midrule
\multicolumn{4}{c}{Entire Dataset} 
  & 0.742 & \underline{0.771} & 0.754 
  & 0.626 & 0.607 & 0.736 & 0.709 
  & \textbf{0.783} \\
\bottomrule[1.8pt]
\end{tabular*}
\caption{\textbf{Comparison of F1-score across modalities and methods on FPRM.} REMIND achieves the best.}
\end{table*}

\begin{table*}[!t]
\centering
\renewcommand{\arraystretch}{1.2}

\label{tab:f1embed-mean}
\begin{tabular*}{\textwidth}{@{\extracolsep{\fill}} c c c c c c c c c c c c }
\toprule[1.8pt]
\multicolumn{4}{c}{\textbf{Missing Scenarios}}
  & \multicolumn{3}{c}{\textbf{Multi‐modal learning methods}}
  & \multicolumn{4}{c}{\textbf{Long Tail}}
  & \textbf{Ours} \\
\cmidrule(lr){1-4}\cmidrule(lr){5-7}\cmidrule(lr){8-11}\cmidrule(lr){12-12}
M1 & M2 & M3 & M4
  & SoftMoE & FuseMoE & FlexMoE
  & FairMixup & Reweigh & GroupDRO & FairBatch
  & REMIND \\
\midrule
\midrule
0 & 0 & 1 & 0
  & 0.641 & 0.643 & \underline{0.646}
  & 0.639 & 0.562 & 0.615 & 0.604
  & \textbf{0.658} \\
1 & 0 & 1 & 0
  & 0.623 & 0.616 & 0.606
  & \textbf{0.697} & 0.634 & \underline{0.676} & 0.627
  & 0.670 \\
0 & 0 & 0 & 1
  & 0.565 & \underline{0.614} & 0.563
  & 0.593 & \textbf{0.618} & 0.559 & 0.566
  & 0.560 \\
0 & 0 & 1 & 1
  & 0.707 & 0.704 & \underline{0.709}
  & 0.710 & \underline{0.720} & \underline{0.720} & 0.700
  & \textbf{0.751} \\
1 & 0 & 1 & 1
  & 0.622 & 0.465 & \underline{0.648}
  & 0.567 & 0.534 & 0.593 & 0.497
  & \textbf{0.673} \\
1 & 1 & 1 & 1
  & 0.776 & 0.768 & 0.750
  & \underline{0.780} & 0.785 & 0.777 & 0.778
  & \textbf{0.803} \\
\midrule
\multicolumn{4}{c}{Entire Dataset}
  & 0.726 & 0.722 & 0.727
  & 0.730 & \underline{0.736} & 0.735 & 0.723
  & \textbf{0.758} \\
\bottomrule[1.8pt]
\end{tabular*}
\caption{\textbf{Comparison of F1-score across modalities and methods on EMBED.} REMIND achieves the best.}
\end{table*}

\begin{table*}[!t]
\centering
\renewcommand{\arraystretch}{1.2}
\label{tab:f1-mimic-mean}
\begin{tabular*}{\textwidth}{@{\extracolsep{\fill}} c c c c c c c c c c c }
\toprule[1.8pt]
\multicolumn{3}{c}{\textbf{Missing Scenarios}} 
  & \multicolumn{3}{c}{\textbf{Multi‐modal learning methods}} 
  & \multicolumn{4}{c}{\textbf{Long Tail}} 
  & \textbf{Ours} \\
\cmidrule(lr){1-3}\cmidrule(lr){4-6}\cmidrule(lr){7-10}\cmidrule(lr){11-11}

M1 & M2 & M3 
  & SoftMoE & FuseMoE & FlexMoE 
  & FairMixup & Reweigh & GroupDRO & FairBatch 
  & REMIND \\
\midrule
\midrule
1 & 0 & 0 
  & \underline{0.625} & 0.583 & 0.611 
  & 0.565 & 0.614 & 0.588 & 0.576 
  & \textbf{0.645} \\
0 & 1 & 0 
  & 0.521 & 0.567 & 0.467 
  & 0.523 & 0.560 & \underline{0.568} & 0.541 
  & \textbf{0.573} \\
1 & 1 & 0 
  & 0.575 & 0.582 & \textbf{0.615} 
  & 0.571 & 0.551 & \underline{0.591} & 0.584 
  & 0.581 \\
0 & 0 & 1 
  & 0.534 & \underline{0.589} & 0.586 
  & \textbf{0.608} & 0.551 & 0.499 & 0.549 
  & 0.550 \\
1 & 0 & 1 
  & \textbf{0.630} & 0.584 & 0.572 
  & 0.594 & 0.589 & \underline{0.627} & 0.590 
  & 0.607 \\
0 & 1 & 1 
  & 0.557 & \textbf{0.598} & 0.546 
  & \underline{0.577} & 0.556 & 0.571 & 0.574 
  & 0.548 \\
1 & 1 & 1 
  & 0.581 & \underline{0.601} & 0.598 
  & \textbf{0.603} & 0.597 & 0.599 & 0.592 
  & \textbf{0.603} \\  
\midrule
\multicolumn{3}{c}{Entire Dataset} 
  & 0.580 & \underline{0.591} & 0.585 
  & 0.587 & 0.581 & 0.590 & 0.586 
  & \textbf{0.594} \\
\bottomrule[1.8pt]
\end{tabular*}
\caption{\textbf{Comparison of F1-score across modalities and methods on MIMIC.} REMIND achieves the best.}
\end{table*}

\begin{table*}[!t]
\centering
\renewcommand{\arraystretch}{1.2}
\label{tab:fprm-f1-std}
\begin{tabular*}{\textwidth}{@{\extracolsep{\fill}} c c c c c c c c c c c c }
\toprule[1.8pt]
\multicolumn{4}{c}{\textbf{Modality}} 
  & \multicolumn{3}{c}{\textbf{Multi‐modal learning methods}} 
  & \multicolumn{4}{c}{\textbf{Long tail}} 
  & \textbf{Ours} \\
\cmidrule(lr){1-4}\cmidrule(lr){5-7}\cmidrule(lr){8-11}\cmidrule(lr){12-12}
M1 & M2 & M3 & M4 
  & SoftMoE & FuseMoE & FlexMoE 
  & FairMixup & Reweigh & GroupDRO & FairBatch 
  & REMIND \\
\midrule
\midrule
1 & 0 & 0 & 0 
  & 0.063 & 0.067 & 0.076 
  & 0.170 & 0.041 & 0.068 & 0.145 
  & 0.069 \\
1 & 1 & 1 & 0 
  & 0.127 & 0.130 & 0.043 
  & 0.077 & 0.371 & 0.072 & 0.124 
  & 0.028 \\
1 & 0 & 0 & 1 
  & 0.141 & 0.025 & 0.014 
  & 0.139 & 0.030 & 0.062 & 0.134 
  & 0.012 \\
1 & 1 & 0 & 1 
  & 0.247 & 0.247 & 0.342 
  & 0.361 & 0.441 & 0.290 & 0.361 
  & 0.000 \\
1 & 1 & 1 & 1 
  & 0.235 & 0.123 & 0.116 
  & 0.173 & 0.221 & 0.200 & 0.171 
  & 0.185 \\
\midrule
\multicolumn{4}{c}{Entire Dataset} 
  & 0.097 & 0.087 & 0.065 
  & 0.056 & 0.167 & 0.110 & 0.093 
  & 0.094 \\
\bottomrule[1.8pt]
\end{tabular*}
\caption{\textbf{Standard Deviation of F1-score across modalities and methods on FPRM.} REMIND achieves the best.}
\end{table*}

\begin{table*}[!t]
\centering
\renewcommand{\arraystretch}{1.2}

\label{tab:f1embed-std}
\begin{tabular*}{\textwidth}{@{\extracolsep{\fill}} c c c c c c c c c c c c }
\toprule[1.8pt]
\multicolumn{4}{c}{\textbf{Missing Scenarios}}
  & \multicolumn{3}{c}{\textbf{Multi‐modal learning methods}}
  & \multicolumn{4}{c}{\textbf{Long Tail}}
  & \textbf{Ours} \\
\cmidrule(lr){1-4}\cmidrule(lr){5-7}\cmidrule(lr){8-11}\cmidrule(lr){12-12}
M1 & M2 & M3 & M4
  & SoftMoE & FuseMoE & FlexMoE
  & FairMixup & Reweigh & GroupDRO & FairBatch
  & REMIND \\
\midrule
\midrule
0 & 0 & 1 & 0
  & 0.015 & 0.021 & 0.028
  & 0.009 & 0.032 & 0.028 & 0.031
  & 0.018 \\
1 & 0 & 1 & 0
  & 0.034 & 0.088 & 0.055
  & 0.031 & 0.037 & 0.137 & 0.095
  & 0.042 \\
0 & 0 & 0 & 1
  & 0.015 & 0.036 & 0.044
  & 0.124 & 0.114 & 0.018 & 0.105
  & 0.073 \\
0 & 0 & 1 & 1
  & 0.003 & 0.002 & 0.004
  & 0.021 & 0.006 & 0.006 & 0.021
  & 0.020 \\
1 & 0 & 1 & 1
  & 0.006 & 0.052 & 0.156
  & 0.124 & 0.105 & 0.066 & 0.046
  & 0.112 \\
1 & 1 & 1 & 1
  & 0.003 & 0.004 & 0.040
  & 0.010 & 0.010 & 0.005 & 0.020
  & 0.004 \\
\midrule
\multicolumn{4}{c}{Entire Dataset}
  & 0.004 & 0.002 & 0.003
  & 0.014 & 0.006 & 0.004 & 0.016
  & 0.003 \\
\bottomrule[1.8pt]
\end{tabular*}
\caption{\textbf{Standard Deviation of F1-score across modalities and methods on EMBED.} REMIND achieves the best.}
\end{table*}

\begin{table*}[!t]
\centering
\renewcommand{\arraystretch}{1.2}
\label{tab:f1-mimic-std}
\begin{tabular*}{\textwidth}{@{\extracolsep{\fill}} c c c c c c c c c c c }
\toprule[1.8pt]
\multicolumn{3}{c}{\textbf{Missing Scenarios}} 
  & \multicolumn{3}{c}{\textbf{Multi‐modal learning methods}} 
  & \multicolumn{4}{c}{\textbf{Long Tail}} 
  & \textbf{Ours} \\
\cmidrule(lr){1-3}\cmidrule(lr){4-6}\cmidrule(lr){7-10}\cmidrule(lr){11-11}
M1 & M2 & M3 
  & SoftMoE & FuseMoE & FlexMoE 
  & FairMixup & Reweigh & GroupDRO & FairBatch 
  & REMIND \\
\midrule
\midrule
1 & 0 & 0  & 0.004 & 0.022 & 0.010 & 0.024 & 0.008 & 0.023 & 0.032 & 0.009 \\
0 & 1 & 0  & 0.004 & 0.014 & 0.011 & 0.020 & 0.005 & 0.026 & 0.027 & 0.008 \\
1 & 1 & 0  & 0.006 & 0.024 & 0.014 & 0.024 & 0.006 & 0.028 & 0.032 & 0.009 \\
0 & 0 & 1  & 0.007 & 0.014 & 0.010 & 0.025 & 0.006 & 0.018 & 0.023 & 0.008 \\
1 & 0 & 1  & 0.010 & 0.019 & 0.015 & 0.022 & 0.003 & 0.032 & 0.034 & 0.009 \\
0 & 1 & 1  & 0.006 & 0.017 & 0.016 & 0.025 & 0.008 & 0.025 & 0.033 & 0.008 \\
1 & 1 & 1  & 0.002 & 0.022 & 0.018 & 0.023 & 0.007 & 0.030 & 0.024 & 0.009 \\  
\midrule
\multicolumn{3}{c}{Entire Dataset} 
  & 0.003 & 0.013 & 0.011 & 0.021 & 0.005 & 0.021 & 0.019 & 0.008\\
\bottomrule[1.8pt]
\end{tabular*}
\caption{\textbf{Standard Deviation of F1-score across modalities and methods on MIMIC.} REMIND achieves the best.}
\end{table*}

\begin{table*}[!t]
\centering
\renewcommand{\arraystretch}{1.2}

\label{tab:embedacc-std}
\begin{tabular*}{\textwidth}{@{\extracolsep{\fill}} c c c c c c c c c c c c }
\toprule[1.8pt]
\multicolumn{4}{c}{\textbf{Missing Scenarios}}
  & \multicolumn{3}{c}{\textbf{Multi‐modal learning methods}}
  & \multicolumn{4}{c}{\textbf{Long Tail}}
  & \textbf{Ours} \\
\cmidrule(lr){1-4}\cmidrule(lr){5-7}\cmidrule(lr){8-11}\cmidrule(lr){12-12}
M1 & M2 & M3 & M4
  & SoftMoE & FuseMoE & FlexMoE
  & FairMixup & Reweigh & GroupDRO & FairBatch
  & REMIND \\
\midrule
\midrule
0 & 0 & 1 & 0
  & 0.008 & 0.005 & 0.024
  & 0.026 & 0.008 & 0.017 & 0.023
  & 0.010 \\
1 & 0 & 1 & 0
  & 0.022 & 0.043 & 0.041
  & 0.006 & 0.027 & 0.043 & 0.011
  & 0.031 \\
0 & 0 & 0 & 1
  & 0.009 & 0.012 & 0.023
  & 0.079 & 0.056 & 0.029 & 0.032
  & 0.020 \\
0 & 0 & 1 & 1
  & 0.002 & 0.004 & 0.002
  & 0.006 & 0.002 & 0.005 & 0.007
  & 0.003 \\
1 & 0 & 1 & 1
  & 0.010 & 0.032 & 0.066
  & 0.052 & 0.031 & 0.010 & 0.028
  & 0.042 \\
1 & 1 & 1 & 1
  & 0.002 & 0.003 & 0.004
  & 0.005 & 0.012 & 0.007 & 0.006
  & 0.003 \\
\midrule
\multicolumn{4}{c}{Total}
  & 0.002 & 0.003 & 0.002
  & 0.007 & 0.005 & 0.006 & 0.009
  & 0.003 \\
\bottomrule[1.8pt]
\end{tabular*}
\caption{\textbf{Standard Deviation of ACC across modalities and methods on EMBED.} REMIND achieves the best.}
\end{table*}

\begin{table*}[!t]
\centering
\renewcommand{\arraystretch}{1.2}

\label{tab:fprm-acc-std}
\begin{tabular*}{\textwidth}{@{\extracolsep{\fill}} c c c c c c c c c c c c }
\toprule[1.8pt]
\multicolumn{4}{c}{\textbf{Modality}} 
  & \multicolumn{3}{c}{\textbf{Multi‐modal learning methods}} 
  & \multicolumn{4}{c}{\textbf{Long tail}} 
  & \textbf{Ours} \\
\cmidrule(lr){1-4}\cmidrule(lr){5-7}\cmidrule(lr){8-11}\cmidrule(lr){12-12}
M1 & M2 & M3 & M4 
  & SoftMoE & FuseMoE & FlexMoE 
  & FairMixup & Reweigh & GroupDRO & FairBatch 
  & REMIND \\
\midrule
\midrule
1 &  &  &  
  & 0.065 & 0.035 & 0.021 
  & 0.038 & 0.039 & 0.064 & 0.038 
  & 0.038 \\
1 & 1 & 1 &  
  & 0.006 & 0.071 & 0.152 
  & 0.043 & 0.122 & 0.072 & 0.022 
  & 0.046 \\
1 &  &  & 1
  & 0.019 & 0.014 & 0.023 
  & 0.031 & 0.020 & 0.012 & 0.012 
  & 0.003 \\
1 & 1 &  & 1
  & 0.144 & 0.144 & 0.250 
  & 0.250 & 0.382 & 0.144 & 0.250 
  & 0.000 \\
1 & 1 & 1 & 1
  & 0.051 & 0.069 & 0.028 
  & 0.071 & 0.074 & 0.156 & 0.045 
  & 0.080 \\
\midrule
\multicolumn{4}{c}{Overall} 
  & 0.028 & 0.018 & 0.017 
  & 0.017 & 0.025 & 0.014 & 0.013 
  & 0.023 \\
\bottomrule[1.8pt]
\end{tabular*}
\caption{\textbf{Standard Deviation of ACC across modalities and methods on FPRM.} REMIND achieves the best.}
\end{table*}

\begin{table*}[!t]
\centering
\renewcommand{\arraystretch}{1.2}

\label{tab:acc-mimic-std}
\begin{tabular*}{\textwidth}{@{\extracolsep{\fill}} c c c c c c c c c c c }
\toprule[1.8pt]
\multicolumn{3}{c}{\textbf{Missing Scenarios}} 
  & \multicolumn{3}{c}{\textbf{Multi‐modal learning methods}} 
  & \multicolumn{4}{c}{\textbf{Long Tail}} 
  & \textbf{Ours} \\
\cmidrule(lr){1-3}\cmidrule(lr){4-6}\cmidrule(lr){7-10}\cmidrule(lr){11-11}
M1 & M2 & M3 
  & SoftMoE & FuseMoE & FlexMoE 
  & FairMixup & Reweigh & GroupDRO & FairBatch 
  & REMIND \\
\midrule
\midrule
1 & 0 & 0  & 0.0161 & 0.0367 & 0.0185 & 0.0347 & 0.0129 & 0.0361 & 0.0617 & 0.0153 \\
0 & 1 & 0  & 0.0140 & 0.0428 & 0.0231 & 0.0426 & 0.0061 & 0.0312 & 0.0242 & 0.0133 \\
1 & 1 & 0  & 0.0104 & 0.0394 & 0.0217 & 0.0425 & 0.0157 & 0.0518 & 0.0434 & 0.0147 \\
0 & 0 & 1  & 0.0095 & 0.0271 & 0.0208 & 0.0254 & 0.0069 & 0.0382 & 0.0357 & 0.0127 \\
1 & 0 & 1  & 0.0106 & 0.0348 & 0.0212 & 0.0426 & 0.0069 & 0.0485 & 0.0495 & 0.0139 \\
0 & 1 & 1  & 0.0132 & 0.0384 & 0.0251 & 0.0294 & 0.0079 & 0.0382 & 0.0286 & 0.0139 \\
1 & 1 & 1  & 0.0141 & 0.0214 & 0.0287 & 0.0381 & 0.0059 & 0.0451 & 0.0482 & 0.0147 \\ 
\midrule
\multicolumn{3}{c}{Total} 
  & 0.0067 & 0.0189 & 0.0171 & 0.0346 & 0.0017 & 0.0349 & 0.0353 & 0.0136\\
\bottomrule[1.8pt]
\end{tabular*}
\caption{\textbf{Standard Deviation of ACC across modalities and methods on MIMIC.} REMIND achieves the best.}
\end{table*}

\subsection{Implementation Details}
Code will be released regardless of acceptance after decision is out. Please search our paper's title by that time.

\subsubsection{FPRM Implementation Details.}
\textbf{Dataset and Preprocessing.} The FPRM dataset contains 3 394 examples from 1 683 patients, each with up to four modalities. All modalities goes through the PatchEmbedding Projection Layer to have the same final number of tokens (16) and embedding dimension (128).
\begin{itemize}
  \item \textbf{M1:} 2D retinal fundus photographs, RGB, resized to $224\times224$, normalized to ImageNet mean/std.
  \item \textbf{M2:} 3D blood-flow volumes, $16$ frames per sample; input permuted to $(C,T,H,W)$ and trilinearly interpolated to $224\times224$ per frame.
  \item \textbf{M3:} 2D oxygen saturation maps, RGB, $224\times224$, normalized similarly to M1.
  \item \textbf{M4:} Tabular metadata are standardized to zero mean and unit variance, then split into $16$ patches of size $\lceil64/16\rceil=4$ for embedding using the patch embedding layer.
\end{itemize}
Modality‐combination distribution:
\[
\begin{array}{ll}
\text{MC1 (M1+M4)} & 59.9\%\\
\text{MC2 (M1 only)} & 17.7\%\\
\text{MC3 (M1+M2+M3+M4)} & 16.2\%\\
\text{MC4 (M1+M2+M3)} & 5.3\%\\
\text{MC5 (M1+M3+M4)} & 0.8\%
\end{array}
\]

\textbf{Model Architecture.} All experiments use a REMIND MoE transformer with 32 experts and one slot per expert, embedding size 128, num of heads 8.
Modality‐specific encoders:
\begin{itemize}
  \item \textbf{PatchEmbeddings} (M4): Linear(D→128) with dropout 0.25, \eg projecting $1\times64\to16\times128$.
  \item \textbf{ResNet2DAdapter} (M1, M3): Pretrained ResNet-50, remove FC head, project 2048→512 via Kaiming-initialized fc.
  \item \textbf{ResNet3DAdapter} (M2): torchvision r3d\_18 (Kinetics-400), remove head, adaptive avg‐pool, project 512→512.
\end{itemize}

\textbf{Training Configuration.}
\begin{itemize}
  \item \textbf{Epochs:} 50; \textbf{Batch size:} 32 per GPU.
  \item \textbf{Optimizer:} AdamW with $\mathrm{lr}=5\times10^{-5}$, $\mathrm{wd}=\mathrm{lr}/10$.

  \item \textbf{Runs:} 3 seeds, reporting mean $\pm$ std.
\end{itemize}

\subsubsection{MIMIC Implementation Details.}
\textbf{Dataset and Preprocessing.} We leverage the MIMIC dataset following~\citep{yun2024flex,thao2024medfuse}, using the clinic text, ICD-9 Code, and Lab test modalities, each sample has up to three modalities. 

\begin{itemize}
  \item \textbf{M1:} clinical text.
  \item \textbf{M2:} labs and vital values.
  \item \textbf{M3:} ICD-9 codes.
\end{itemize}
During preprocessing, we randomly masked some existing modalities for each record to construct comprehensive modality combinations, and the modality-combination's distribution is:
\[
\begin{array}{ll}
\text{MC1 (M1 only)} & 4.8\%\\
\text{MC2 (M2 only)} & 5.2\%\\
\text{MC3 (M1+M2)} & 14.3\%\\
\text{MC4 (M3 only)} & 5.2\%\\
\text{MC5 (M1+M3)} & 15.5\%\\
\text{MC6 (M2+M3)} & 14.7\%\\
\text{MC7 (M1+M2+M3)} & 40.3\%
\end{array}
\]

\textbf{Model Architecture.} All experiments use a REMIND MoE transformer with 128 experts and one slot per expert, embedding size 768, num of heads 8. We treat each modality as text input, which are then encoded by modality-specific T5-base encoders~\citep{raffel2020exploring}.

\textbf{Training Configuration.}
\begin{itemize}
  \item \textbf{Epochs:} Max 20; \textbf{Batch size:} 32 per GPU.
  \item \textbf{Optimizer:} AdamW with $\mathrm{lr}=1\times10^{-4}$, $\mathrm{wd}=5\times10^{-5}$.
  \item \textbf{Runs:} 3 seeds, reporting mean $\pm$ std.
\end{itemize}

\textbf{EMBED Implementation Details.}

\textbf{Dataset and Preprocessing.} We leverage the EMBED dataset following~\citep{wu2025dynamic}, and each sample has up to four modalities.

\begin{itemize}
  \item \textbf{M1:} C-View synthetic Cranio-Caudal view image, resized to $224\times224$, normalized to ImageNet mean/std.
  \item \textbf{M2:} C-View synthetic 
Medio-Lateral Oblique view image, resized to $224\times224$, normalized to ImageNet mean/std.
  \item \textbf{M3:} full-field digital mammography Cranio-Caudal view image, resized to $224\times224$, normalized to ImageNet mean/std.
  \item \textbf{M4:} full-field digital mammography Medio-Lateral Oblique view image, resized to $224\times224$, normalized to ImageNet mean/std.
\end{itemize}

Modality-combination distribution:
\[
\begin{array}{ll}
\text{MC1 (M3 only)} & 3.0\%\\
\text{MC2 (M1+M3)} & 0.6\%\\
\text{MC3 (M4 only)} & 1.2\%\\
\text{MC4 (M3+M4)} & 57.8\%\\
\text{MC5 (M1+M3+M4)} & 0.3\%\\
\text{MC6 (M1+M2+M3+M4)} & 37.1\%
\end{array}
\]

\textbf{Model Architecture.} All experiments use a REMIND MoE transformer with 128 experts and one slot per expert, embedding size 768, num of heads 8. Each modality are firstly projected by modality-specific ViT-base encoders~\citep{dosovitskiy2020image}.

\textbf{Training Configuration.}
\begin{itemize}
  \item \textbf{Epochs:} Max 20; \textbf{Batch size:} For FairMixup, 16 per GPU because it takes larger GPU memory; For others, 32 per GPU.

  \item \textbf{Optimizer:} AdamW with $\mathrm{lr}=2\times10^{-5}$ for FairMixup and $\mathrm{lr}=5\times10^{-5}$ for others, $\mathrm{wd}=5\times10^{-5}$.

  \item \textbf{Runs:} 3 seeds, reporting mean $\pm$std.
\end{itemize}

Main experiments are conducted on NVIDIA A100-SXM4-80GB GPUs. All datasets use FocalLoss($\gamma$=2), and a scheduler of StepLR($\mathrm{step\_size}=5$, $\gamma=0.1$).

\subsection{Discussion Q3/Table 5 Experiment Explained}

Extreme Missingness Robustness Analysis. To evaluate robustness under extreme missing scenarios, we systematically simulate high missingness rates for individual modalities. For each modality $M_i$ (where $i \in {1,2,3,4}$), we artificially remove it from 80\% of samples, creating two distinct groups: the "Non-Missing" group containing the remaining 20\% of samples that retain $M_i$, and the "Missing" group comprising the 80\% of samples without $M_i$. We then evaluate both baseline methods and our approach on: (1) overall performance across all samples, (2) performance on the Non-Missing group, and (3) performance on the Missing group. This design allows us to assess whether methods can effectively utilize sparse modalities when they are available while maintaining robust performance when they are absent.
The results in Table~\ref{tab:extreme_missing} demonstrate that our method exhibits superior robustness compared to baselines. Each row represents the experiment where modality $M_i$ is made 80\% missing. Notably, our approach maintains more consistent performance across both groups, with smaller performance gaps between Non-Missing and Missing groups, indicating better adaptation to varying modality availability patterns.
\begin{table*}[t]
\centering
\small
\resizebox{\textwidth}{!}{%
\begin{tabular}{cccccccccccc}
\toprule[1.8pt]
\multirow{2}{*}{Scenario} &
\multicolumn{4}{c}{\textbf{Modality presence (1 = observed)}} &
\multicolumn{2}{c}{\textbf{DRO step size $\eta$}} &
\multicolumn{3}{c}{\textbf{\# fusion experts $E$}} &
\multicolumn{2}{c}{\textbf{Initialization}} \\
\cmidrule(r){2-5}\cmidrule(r){6-7}\cmidrule(r){8-10}\cmidrule(r){11-12}
& M1 & M2 & M3 & M4
& 0.1 & 0.5
& 32 & 64 & 128
& zero emb.
& sinus-resid.\\
\midrule
\midrule
S1 &   &   & 1 &   & 74.6 & 75.6 & 73.8 & 74.1 & 74.6 & 73.3 & 73.6 \\
S2 & 1 &   & 1 &   & 81.7 & 77.4 & 82.8 & 81.7 & 79.9 & 84.9 & 77.4 \\
S3 &   &   &   & 1 & 75.7 & 75.1 & 70.5 & 78.0 & 74.6 & 75.1 & 76.9 \\
S4 &   &   & 1 & 1 & 79.2 & 79.1 & 78.8 & 79.2 & 79.5 & 79.1 & 79.3 \\
S5 & 1 &   & 1 & 1 & 74.5 & 76.4 & 72.7 & 69.1 & 74.0 & 69.1 & 65.5 \\
S6 & 1 & 1 & 1 & 1 & 84.0 & 84.0 & 84.0 & 84.4 & 84.0 & 84.1 & 83.5 \\
\midrule
Avg &   &   &   &   & 80.5 & 80.4 & 80.2 & 80.6 & 80.7 & 80.5 & 80.3 \\
\bottomrule[1.8pt]
\end{tabular}%
}
\caption{\textbf{Hyperparameter sensitivity of REMIND on the EMBED dataset across six representative missingness scenarios.} Columns on the left indicate which modalities (M1--M4) are observed in each scenario (1 = present, blank = missing). We vary the group-DRO step size $\eta$, the number of fusion experts $E$, and initialization schemes for imputation embeddings and group-specific components. The last row reports the average performance (macro AUROC) over all scenarios.}
\label{tab:hparam_full}
\end{table*}
\subsection{Hyperparameter Sensitivity on EMBED}
\label{app:hparam_embed}

We study the sensitivity of REMIND to several key hyperparameters on the EMBED dataset under different missing-modality patterns. Table~\ref{tab:hparam_full} reports results for six representative missingness scenarios (rows S1--S6), where the binary block on the left indicates which of the four modalities (M1--M4) are observed in each scenario (1 = present, blank = missing). For each scenario we vary:
\begin{itemize}
    \item the group-DRO step size $\eta \in \{0.1, 0.5\}$,
    \item the number of fusion experts $E \in \{32, 64, 128\}$,
    \item imputation embedding and initialization schemes for the group-specific router (disabling learnable imputation embeddings, ``zero emb'', and sinusoidal initialization of residual routing matrices, ``sinus-resid'').
\end{itemize}

Across scenarios, the average accuracy (last row) remains tightly clustered between $80.2\%$ and $80.7\%$, indicating that REMIND is fairly insensitive to these hyperparameters. Increasing the number of experts from $32$ to $64$ or $128$ yields small but consistent gains (from $80.2\%$ to $80.6\%$ and $80.7\%$), while changing the DRO step size from $\eta=0.1$ to $\eta=0.5$ has negligible effect (from $80.5\%$ to $80.4\%$ on average). The routing initialization choices and whether to use a learnable imputation embedding and  also lead to similar overall performance ($80.3\%$ and \ $80.5\%$).
\begin{table}[h]
\small
\centering
\resizebox{0.5\textwidth}{!}{%
\begin{tabular}{l l l l}
\toprule[1.8pt]
\bf Method           & \bf Parameters                                                      & \bf Step per Sec &\bf Flops        \\
\midrule
\midrule
SoftMoE & 958 M & 0.41 & 146.7 G\\
REMIND & 960 M & 0.42 & 146.7 G \\
REMIND (with group-specific routing) & 960 M & 0.68 & 146.9 G\\
\bottomrule[1.8pt]
\end{tabular}
}
\caption{\textbf{Computation costs.}}
\label{tab: computation cost}
\end{table}

REMIND uses an exponentiated-gradient update for the group weights $\{\lambda_k\}$ in the group-DRO objective, with a sharpness parameter $\gamma$ controlling how aggressively the optimizer focuses on high-loss groups. On the EMBED dataset, we sweep $\gamma \in \{0.5, 0.1, 0.02\}$ and obtain overall accuracies of $80.4\%$, $80.5\%$, and $80.7\%$, respectively. The best performance is achieved with the smallest value $\gamma = 0.02$ (used in our main experiments), suggesting that a milder sharpness that does not over-concentrate on the worst groups yields slightly better overall robustness, while performance remains stable across this range.

\subsection{Computation Cost}
Here we report the computation cost of REMIND and a MoE baseline method. The two methods are implemented with the same encoder, and the same configuration of the expert layer in the Soft MoE block. The only difference is that REMIND is built on distributionally robust learning framework, and has modality-specific routing matrices in the second stage training. We compare the parameter size, training time (iteration per second), and the FLOPs per sample in Table~\ref{tab: computation cost}. The training and inference time is per step, where the batch size is 32.

\subsection{Limitations}
Our study has several limitations that suggest promising directions for future work. 
Our empirical evaluation focuses on clinical datasets with three to four modalities constrained by the availability of public high-modality benchmarks, while our proposed framework is generalizable and scalable to even more modalities.
REMIND only adds lightweight group-specific residual routing and embedding parameters based on a shared expert pool with small computational overhead. We also primarily investigate medical classification tasks and extension to segmentation and other multi-task learning is a promising extension. We hope that REMIND provide a useful foundation for building robust and scalable multi-modal systems under realistic high-modality missingness.